\documentclass[a4paper]{cas-sc}

\usepackage[authoryear]{natbib}
\usepackage{rotating} 
\usepackage{multirow} 
\usepackage{bm}
\usepackage{wrapfig}
\usepackage{caption}  
\usepackage{adjustbox}
\usepackage{hyperref}

\def\tsc#1{\csdef{#1}{\textsc{\lowercase{#1}}\xspace}}
\tsc{WGM}
\tsc{QE}

\begin{document}
\let\WriteBookmarks\relax
\def\floatpagepagefraction{1}
\def\textpagefraction{.001}

\shorttitle{Medical Image Analysis (2025)} 

\shortauthors{Yuan Zhang et~al.}   

\title [mode = title]{PathFL: Multi-Alignment Federated Learning for Pathology Image Segmentation} 

\tnotemark[1] 


\author[1]{Yuan Zhang}[type=editor,
                       orcid=0000-0002-0762-7514]
\ead{230219180@seu.edu.cn}

\author[2]{Feng Chen}[orcid=0000-0002-2699-7190]%
\ead{fengchen@njmu.edu.cn} 

\author[1]{Yaolei Qi}[orcid=0000-0002-8531-7386]%
\ead{1024443299@qq.com}   
  
\author[1]{Guanyu Yang}[orcid=0000-0003-3704-1722]%
\ead{yang.list@seu.edu.cn}  
\cormark[1]

\author[3]{Huazhu Fu}[orcid=0000-0002-9702-5524]%
\ead{hzfu@ieee.org} 
\cormark[1]
\cortext[cor1]{Corresponding author}

\affiliation[1]{organization={Key Laboratory of New Generation Artificial Intelligence Technology and Its Interdisciplinary Applications (Southeast University), Ministry of Education},
            addressline={No.2, Sipai Lou, Xuanwu District}, 
            city={Nanjing},
            postcode={210096}, 
            country={China}}

\affiliation[2]{organization={Department of Biostatistics, Center for Global Health, School of Public Health, Nanjing Medical University},
            city={Nanjing},
            postcode={211166}, 
            country={China}}

\affiliation[3]{organization={Institute of High-Performance Computing, Agency for Science, Technology and Research},
            postcode={138632}, 
            country={Singapore}}
            
\begin{abstract}
Pathology image segmentation across multiple centers encounters significant challenges due to diverse sources of heterogeneity including imaging modalities, organs, and scanning equipment, whose variability brings representation bias and impedes the development of generalizable segmentation models. In this paper, we propose PathFL, a novel multi-alignment Federated Learning framework for pathology image segmentation that addresses these challenges through three-level alignment strategies of image, feature, and model aggregation. Firstly, at the image level, a collaborative style enhancement module aligns and diversifies local data by facilitating style information exchange across clients. Secondly, at the feature level, an adaptive feature alignment module ensures implicit alignment in the representation space by infusing local features with global insights, promoting consistency across heterogeneous client features learning. Finally, at the model aggregation level, a stratified similarity aggregation strategy hierarchically aligns and aggregates models on the server, using layer-specific similarity to account for client discrepancies and enhance global generalization. Comprehensive evaluations on four sets of heterogeneous pathology image datasets, encompassing cross-source, cross-modality, cross-organ, and cross-scanner variations, validate the effectiveness of our PathFL in achieving better performance and robustness against data heterogeneity. {The code is available at \href{https://github.com/yuanzhang7/PathFL}{https://github.com/yuanzhang7/PathFL}.}
\end{abstract}


\begin{keywords}
Pathology image \sep Federated learning \sep Segmentation \sep Heterogeneity
\end{keywords}

\ExplSyntaxOn
\keys_set:nn { stm / mktitle } { nologo }
\ExplSyntaxOff

\maketitle

\section{Introduction}
Pathology image segmentation plays a crucial role in clinical diagnostics, enabling the accurate identification and classification of various disease markers \citep{wang2019pathology}. In particular, accurate segmentation of nuclear, cellular, and tissue changes serves as a pivotal indicator for a pathologist in the evaluation of cancer malignancy degree and the identification of tumor regions \citep{irshad2013pathreview}. However, the scarcity of annotated data in individual medical centers hinders its performance and generalizability. Furthermore, data sharing between different hospitals faces legal and practical hurdles in realistic clinical scenarios. Thus, the sensitive nature of pathology images, intertwined with patient privacy and ethical concerns necessitates the adoption of multi-center collaborative training under privacy preservation techniques. {The broader challenges of data ownership, regulatory restrictions, and institutional collaboration barriers remain significant obstacles in healthcare \citep{van2014systematic}. Many pathology datasets hold commercial and proprietary value, making institutions reluctant to share them freely \citep{rieke2020future}. Additionally, clinical pathology data are subject to strict governance such as the US Health Insurance Portability and Accountability Act \citep{cheng2006HIPPA} and the EU General Data Protection Regulation \citep{regulation2018general}, with concerns over data security, stewardship, and intellectual property rights. FL provides a viable pathway to overcome these challenges by allowing institutions to benefit from collaborative learning while maintaining control over their proprietary data. It also reduces the need for centralized data storage, mitigating risks associated with information leakage and data breaches. Moreover, FL supports large-scale, multi-institutional collaborations, allowing distributed learning across different centers while respecting data privacy and regulatory constraints \citep{sheller2020federated}}, thereby preserving confidentiality while promoting collaborative advancements in pathology image segmentation.

\begin{wrapfigure}{r}{0.46\textwidth}
    \vspace{-16pt} 
    \centering
    \includegraphics[width=0.46\textwidth]{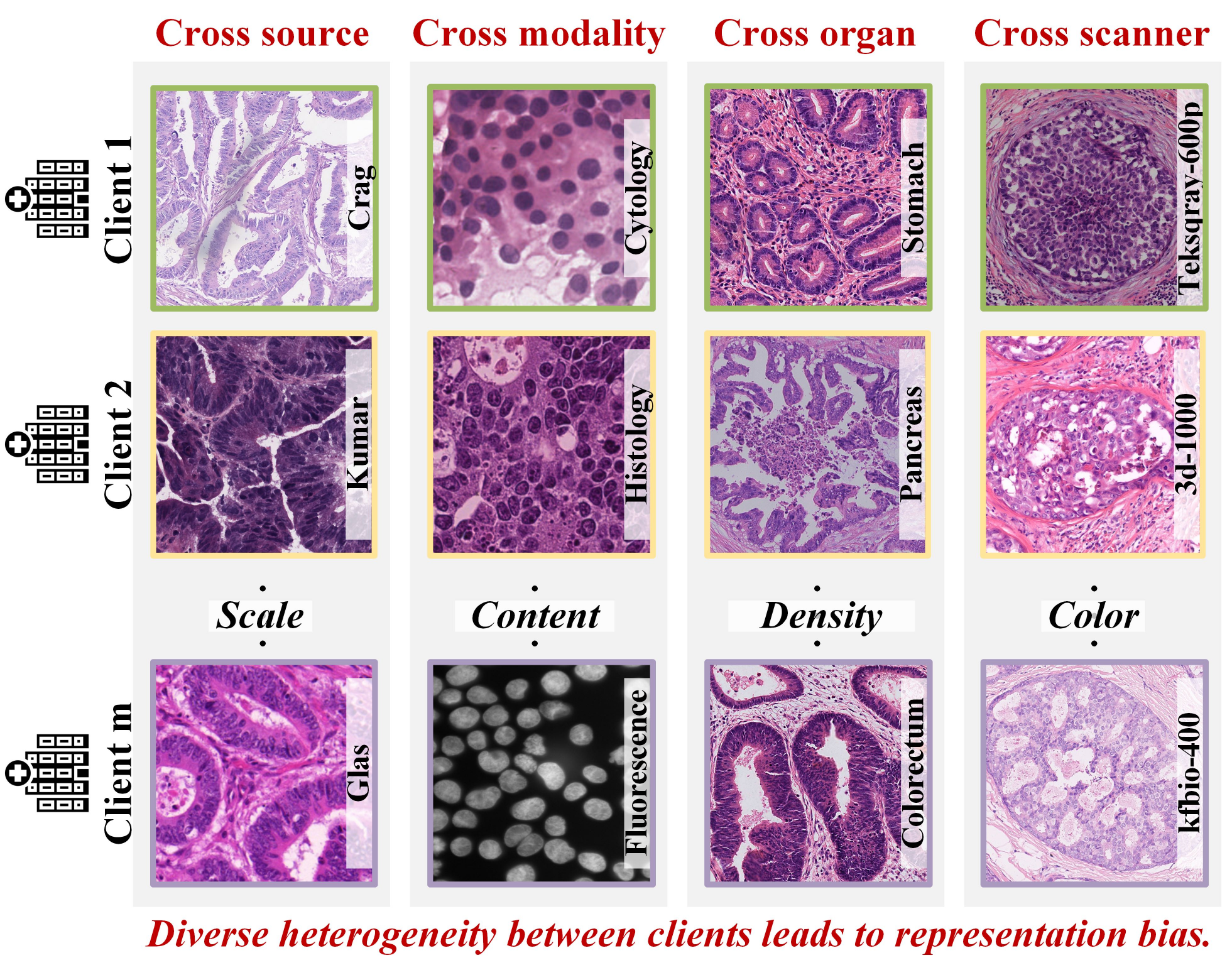}
    \caption{Illustration of heterogeneity for pathology images. The diverse sources of heterogeneity in multi-center pathology image segmentation highlight the significant representation bias across centers in four key aspects: cross-source, cross-modality, cross-organ, and cross-scanner.}
    \label{fig_1_challenge}
    \vspace{-16pt} 
\end{wrapfigure}

The extensive heterogeneity inherent in pathology images from multiple centers presents a significant challenge for FL. Unlike gray-scale medical images, pathology images exhibit pronounced heterogeneity due to richer color variations, encompassing disparities in texture, scale, content, and density. These variations arise from multiple factors, including differences in diverse organs, imaging modalities, staining techniques, and image resolution, as depicted in Fig.~\ref{fig_1_challenge}. To comprehensively evaluate heterogeneity challenges in multi-center pathology segmentation, we constructed four datasets representing \textbf{cross-source}, \textbf{cross-modality}, \textbf{cross-organ}, and \textbf{cross-scanner} scenarios. The high diversity of morphological features amplifies representation bias, introducing substantial inductive bias that exacerbates divergence among local model updates and diminishes global generalization. 

\begin{wrapfigure}{l}{0.45\textwidth}
    \vspace{-15pt} 
    \centering
    \includegraphics[width=0.46\textwidth]{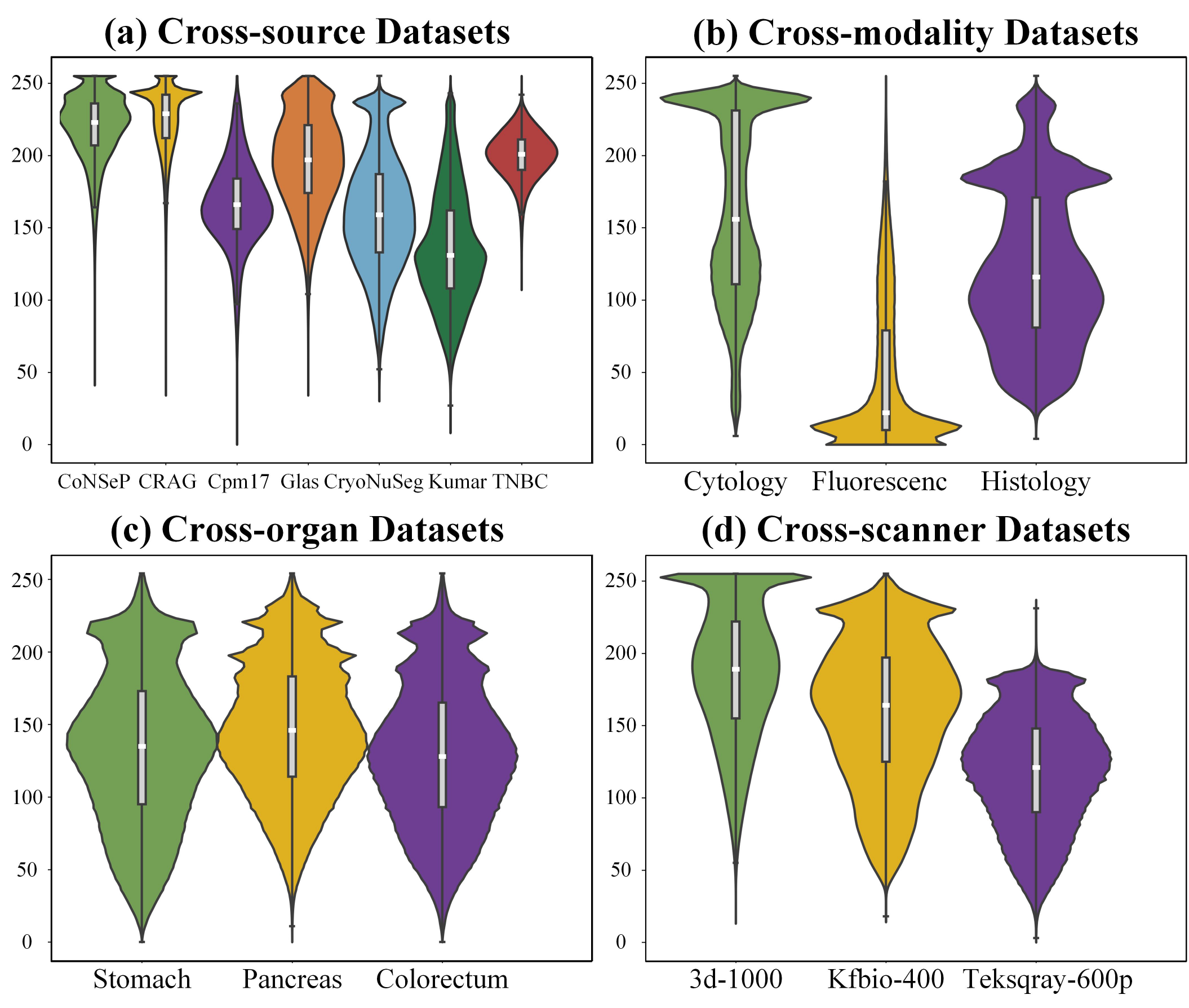}
    \caption{Intensity distributions of four sets of pathology segmentation datasets.  Violin plots reveal both intra-group and inter-group distribution differences, highlighting diverse heterogeneity.}
    \label{fig_3_distribution}
    \vspace{-15pt} 
\end{wrapfigure}

Although the effectiveness of FL for pathology segmentation has been demonstrated by the prior studies \citep{schoenpflugreview}, it comes with limitations. There are three key factors that are greatly influenced by the heterogeneity: input images, latent features, and aggregation mechanisms. 
\textbf{First,} at the image level, the diversity of distribution differences in heterogeneous data from various sources is illustrated in Fig.~\ref{fig_3_distribution}. Variations in intensity distributions and visual characteristics of input images across different centers often lead to inconsistent learning patterns among clients, which may overfit to specific data distributions, leading to poor performance on unseen datasets. 
\textbf{Second,} at the feature level, the features extracted from heterogeneous inputs can vary significantly, leading to representation bias, where the global model disproportionately favors features from dominant or similar clients while neglecting essential features from distinct or smaller datasets. As a result, the model may not adequately capture the full range of characteristics present in the overall data population. 
\textbf{Lastly,} this heterogeneity amplifies imbalances in the aggregation process, as updates from clients with significantly divergent data can conflict or dominate, resulting in suboptimal updates to the global model and poor generalization performance. Most existing methods focus on only one aspect of heterogeneity, without integrating a comprehensive design that considers all three levels of heterogeneity simultaneously.

In this paper, we propose a multi-alignment framework to address heterogeneity by incorporating image-level style enhancement to diversify input distributions, representation-level feature alignment to unify latent feature spaces across clients, and model-level stratified similarity aggregation to balance and refine global model updates. We posit that by effectively leveraging statistical insights, the multi-layer exchange of shallow texture details and deep semantic features in segmentation tasks can enhance the model's representational capability \citep{zhang2018exfuse}. The richness of color in pathology images is fundamentally different from other medical images, and multi-center pathology data is naturally and specifically richer in stylistic information, yet this is often overlooked \citep{yamashita2021learning}. This suggests that the structural and shape characteristics of cells captured in pathology images from different centers are fundamentally alike, even though they may be influenced by varying backgrounds, colors, and imaging conditions \citep{hoque2024stain}. The underlying hypothesis is that cross-center pathology images exhibit similarities in cellular morphology. The style information of an image can be effectively characterized by statistical features, specifically the mean and variance \citep{huang2017adain,li2021feature}. The mean captures the overall brightness level of the image, while the variance provides insights into contrast and texture details. (1) By exchanging stylistic statistical information at the image level, we propose the \textbf{collaborative style enhancement} module to achieve cross-client style exchange, which not only enhances the model's adaptability to diverse data distributions but also aligns data across different centers. (2) By exchanging global mean and variance at the feature level, we propose the \textbf{adaptive feature alignment} module which allows local models to explore the feature space more thoroughly, aligning the latent features learned by different clients. This collaborative mechanism helps unify the representations from multiple perspectives, reducing negative interference between clients. As a result, the global model's generalization performance is enhanced, with the capability to extract more universal features that improve segmentation accuracy. Moreover, only transmitting mean and variance without disclosing the actual feature content or raw data ensures compatibility with the privacy-preserving mechanism of federated learning, while providing a simple and efficient approach for aligning feature distributions across clients. (3) By evaluating layer-wise similarity, we proposed the \textbf{stratified similarity aggregation} strategy allows each client greater flexibility during model updates, aligns model parameters hierarchically at the server, effectively managing client discrepancies and improving generalization.

Our primary contributions are as follows:
\begin{itemize}
\item Framework: We propose PathFL, a multi-alignment framework to address cross-center heterogeneity in pathology image segmentation. To our knowledge, PathFL offers the first unified FL solution that aligns three levels of image, feature, and aggregation.
\item Module design: At the image level, a collaborative style enhancement module diversifies local images by style exchange across clients, thereby enhancing the diversity of input and aligning them across different centers. At the feature level, an adaptive feature alignment module ensures implicit alignment within the latent feature space, promoting consistency of representation. At the aggregation level, a stratified similarity aggregation strategy hierarchically aligns and aggregates models on the server, improving global generalization.
\item Theoretical analysis: We provide a theoretical analysis of our algorithm, elucidating how cross-center style enhancement enhances model performance. Specifically, we demonstrate that exchanging statistical features, such as mean and variance, facilitates effective knowledge transfer, thereby improving model efficacy.
\item Extensive validation: We conducted comprehensive experiments on four distinct pathology image segmentation datasets, exploring various types of heterogeneity commonly encountered. This demonstrates the effectiveness of our approach in handling cross-center heterogeneity.
\end{itemize}

\section{Related Works}
\subsection{Federated learning for pathology images}
Collaborative training across centers risks privacy breaches and faces legal and logistical challenges in clinical practice, making data sharing between institutions impractical \citep{guan2024federatedmedicalsurvey}. FL is increasingly favored for pathology image analysis, offering a solution to leverage multi-institutional datasets without centralizing data, thus preserving privacy and addressing interoperability issues \citep{schoenpflugreview}. Current Research on FL in pathology primarily focuses on three tasks: 1) Classification task. Many studies on pathology image classification \citep{shi2024vila} explore fairness \citep{hosseini2023proportionally}, privacy \citep{lu2022histofl}, and communication cost \citep{lin2023federatedhypernet}, etc. 2) Staining normalization task. In pathological imaging, inconsistencies introduced by varying staining protocols increase the complexity of automated processing, driving considerable attention in the field of research \citep{wagner2022federated, shen2022federated}. 3) Segmentation task. Due to the scarcity of precise annotations for pathology images, research specifically focused on pathology image segmentation remains relatively limited. \citet{ma2024model} explores the issue of model heterogeneity in medical image segmentation, while \citet{zhang2024federatedconnitue} investigates the problem of catastrophic forgetting in medical image models. Additionally, \citet{pan2024feddp} emphasizes the importance of privacy protection and \citet{chen2024think} evaluates the uncertainty through active learning. Given the significant clinical importance of pathology images, many FL methods for medical images incorporate pathology datasets to extend the applicability of their methods. However, these works often overlook the extensive and unique heterogeneity present in pathology images. In contrast, our approach is the first to conduct comprehensive experiments and validations specifically addressing the diverse heterogeneity in pathology image segmentation.

\subsection{Heterogeneity in federated learning}
Existing research in FL for pathology image segmentation has primarily addressed heterogeneity by focusing on isolated factors, without adequately considering the integration of input image, feature representation, and model aggregation \citep{qi2024model}. Firstly, HarmoFL \citep{jiang2022harmofl} addresses input image heterogeneity by normalizing images' frequency domain characteristics on pathology images. Secondly, FedBN \citep{li2021fedbn} targets batch normalization layers to mitigate feature distribution discrepancies, and FedFA \citep{zhou2023fedfa} augments local feature statistics to address feature shifts. Thirdly, numerous methods primarily focus on improving model aggregation mechanisms to mitigate inter-client variations. FedAvg \citep{McMahan2017fedavg} applies the average aggregation of clients, which struggles with non-IID data distributions. FedProx \citep{li2020fedProx} introduces a proximal regularization term to constrain local model divergence, while HistoFL \citep{lu2022histofl} injects random noise into model weights to preserve privacy. FedHeal \citep{chen2024fair} refines aggregation by pruning unimportant parameters and reducing variance between local and global models. Despite notable advancements, existing methods remain constrained by their narrow scope, typically addressing only one dimension of heterogeneity while overlooking the intricate interdependence between input, feature representation, and model aggregation. These three factors are not only directly influenced by heterogeneity but also mutually interact, compounding their impact and imposing significant constraints on performance. Thus, we propose a comprehensive multi-alignment framework that systematically addresses heterogeneity across input, feature, and model aggregation levels, thereby substantially enhancing federated learning robustness and generalization capabilities.

\section{Methodology}
\begin{figure*}[htp]
\centering
\includegraphics[width=\textwidth]{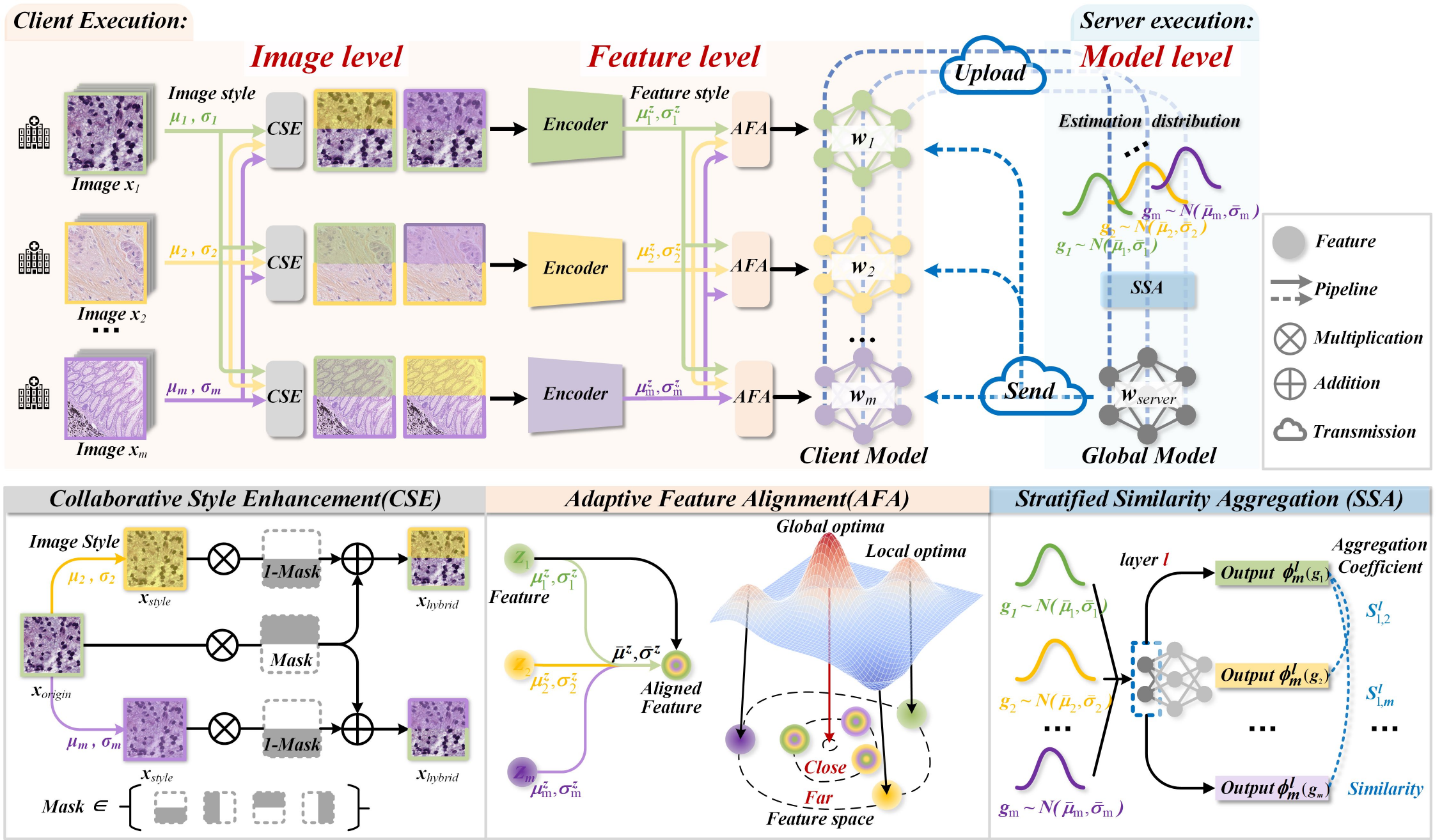}
\caption{Framework illustration of our PathFL, including three key modules: (1) Image-level collaborative style enhancement (CSE), which enriches the stylistic differences in images across clients by transferring essential statistical features; (2) Feature-level adaptive feature alignment (AFA), enabling the exchange of feature distributions between client models to improve alignment and mitigate heterogeneity; and (3) Model-level stratified similarity aggregation (SSA), which aligns and aggregates network layer outputs across clients to enhance model robustness and generalization.}
\label{fig_2_framework}
\end{figure*}

In this section, we first outline the preliminaries and provide a detailed formulation of the problem. We then introduce our approach to stylistic information exchange, leveraging the principles of Vicinal Risk Minimization (VRM) to mitigate the impact of cross-client data heterogeneity. Specifically, we present three modules: the collaborative style enhancement module, adaptive feature alignment module, and stratified-layer aggregation strategy. An overview of our framework PathFL is depicted in Fig. \ref{fig_2_framework}.

\subsection{Problem Formulation}
FL involves training a server model across $M$ distinct clients without transferring data directly. Each client retains its own distribution $D_m= \{(\bm{x}_i, \bm{y}_i)\}_{i=1}^{n_m} $ ($m = 1,\cdots,M$), where $(\bm{x_i}, \bm{y_i})$  represent the image-label pairs and $n_m$ is the number of samples in client $m$. Due to the inherent variability in pathology images across source, modality, organ, and scanner, the samples in different clients exhibit \textit{non-IID} (non-independent and identically distributed) characteristics, leading to significant heterogeneity across clients. More formally, for any two clients $m$ and $\tilde{m}$, the expectation of their local loss functions $f_i(\bm{w}; \bm{x}_i)$ generally differs:
\begin{equation}
\mathbb{E}_{\bm{x}_i \sim \mathcal{D}_m} [f_i(\bm{w}; \bm{x}_i)] \neq \mathbb{E}_{\mathbf{x}_i \sim \mathcal{D}_{\tilde{m}}} [f_i(\bm{w}; \bm{x}_i)],
\label{eq3}
\end{equation}
where $\bm{w}$ represents the model parameters. As a result, local datasets from different clients may not be representative of the global data distribution, i.e., 
\begin{equation}
\mathbb{E}_{\bm{x}_i \sim \mathcal{D}_k} [f_i(\bm{w}; \bm{x}_i)] \neq F_{\text{global}}(\bm{w}),
\label{eq4}
\end{equation}
which poses challenges for model generalization because the aggregated global model may not effectively capture the diverse distributions present across the clients. Furthermore, this heterogeneity can lead to imbalanced aggregation of the global model, resulting in degraded performance and compromised stability of the deep learning algorithm. 

Theoretically, as the number of participating clients $M$ increases and data heterogeneity diminishes, the global expectation $F_{\text{global}}(\bm{w})$ can asymptotically approach the weighted average of client-specific expectations. In the limit of infinite clients with sufficiently similar data distributions, this global model approximates the true data distribution:
\begin{equation}
\lim_{M \to \infty} \frac{1}{M} \sum_{m=1}^{M} \mathbb{E}_{\bm{x}_i \sim \bm{D}_m} [f(\bm{w}; \bm{x}_i)] \approx \mathbb{E}_{\bm{x}_i \sim \mathcal{D}_{\text{global}}} [f(\bm{w}; \bm{x}_i)],
\label{eq5}
\end{equation}
where $\mathcal{D}_{\text{global}}$ represents the cumulative distribution of data from all clients.

\subsection{Theory: Vicinal risk minimization}
Vicinal Risk Minimization (VRM) \citep{chapelle2000vicinal} improves the approximation of the local distribution by introducing ``vicinal distributions'' of the data, which is achieved by incorporating statistical style information, such as mean and variance, from other clients to expand each client's local data distribution, forming a closer approximation of the global distribution $\mathcal{D}_{\text{global}}$. VRM introduces flexibility into the optimization problem by broadening the scope of the local distribution. When clients receive style information from others, they effectively extend their data distribution to encompass patterns and features that were previously absent or unseen. Thus, each client can approximate a distribution $\mathcal{V}(\mathcal{D}_m)$ that not only captures its unique data characteristics but also introduces variations aligned with the broader diverse patterns. This approximation provides robustness by mitigating local biases, as each client is less likely to overfit to its own potentially skewed data, resulting in improved model stability and performance across the federation. Specifically, the objective function of VRM is:

\begin{equation} 
F_{\text{VRM}}(\bm{w}) = \mathbb{E}_{\bm{x} \sim \mathcal{V}(\mathcal{D})} \left[ f(\bm{w}; \bm{x}) \right], 
\end{equation}
where $\mathcal{V}(\mathcal{D})$ represents the vicinal distribution generated around the true data distribution $\mathcal{D}$. In this manner, VRM not only accounts for the information of local samples but also integrates the variability of their neighboring samples, thereby capturing potential data features more comprehensively. By collaboratively aligning stylistic patterns across clients, our approach enriches sample representations, enabling the learning process to effectively transcend local data limitations and capture more generalized feature representations.

Mathematically, VRM can be viewed as an effective approximation of the global distribution. Specifically, as the number of clients $M$ approaches infinity, the VRM theory can be expressed as:
\begin{equation}
F_{\text{VRM}}(\bm{w}) \approx F_{\text{global}}(\bm{w}),
\end{equation}
which indicates VRM is better positioned to approximate the true data distribution, particularly in scenarios where the data distributions exhibit significant heterogeneity. By mitigating the discrepancy between local and global expectations through a VRM-based approach, we effectively enhance model generalization across diverse client distributions. In the context of federated learning, where significant variability in client data distributions is prevalent, the proposed vicinal distribution methodology offers a more precise approximation of local data characteristics, consequently improving model generalization and robustness while systematically addressing the inherent challenges of \textit{non-IID} data across heterogeneous client environments.

\subsection{Collaborative Style Enhancement}
Building upon the principles of VRM, we propose the collaborative style enhancement module that capitalizes on the variability introduced by neighboring distributions within the federated learning framework. For pathology image segmentation tasks, image-level style exchange is crucial as it contains the most detailed texture information necessary for effective analysis \citep{tang2024domain}. By aligning style statistics between clients, we enrich local models with broader stylistic features. By integrating vicinal risk minimization (VRM) theory, our approach significantly enhances model generalization across heterogeneous datasets, effectively incorporating local variability to achieve a more comprehensive risk estimation. This methodological integration not only harmonizes with VRM principles but also fosters a robust model capable of seamlessly capturing diverse image representations in environments characterized by substantial data distribution variations. {To quantify and align data distributions across different clients, we compute statistical features—specifically, the means and standard deviations—at two levels: (1) the image level, characterizing the global intensity distribution of pixel values within each client (Equations \ref{eq:image_mu}, \ref{eq:image_std}), and (2) the feature level, capturing the statistical properties of high-dimensional representations (Equations \ref{eq:feature_mu}, \ref{eq:feature_std}).}

For client $m$ at the $t$-th training epoch in federated learning, we compute the mean $\mu_m^{(t)}$ and standard deviation $\sigma_m^{(t)}$ at each batch as the local stylistic statistics:
\begin{equation}
\mu_m^{(t)} = \frac{1}{UV} \sum_{u=1}^{U} \sum_{v=1}^{V} \bm{x}_m^{u,v (t)} \in \mathbb{R},
\label{eq:image_mu}
\end{equation}
\begin{equation}
\sigma_m^{(t)} = \sqrt{\frac{1}{UV} \sum_{u=1}^{U} \sum_{v=1}^{V} \left(\bm{x}_m^{u,v(t)} - \mu_m^{(t)} \right)^2} \in \mathbb{R}.
\label{eq:image_std}
\end{equation}
Here, $U$ and $V$ represent the height and width of the images, and $\bm{x}_m^{u,v(t)}$ denotes the pixel intensity at position $(u, v)$ for client $m$ at epoch $t$. To facilitate style transfer, we aggregate these statistics from all clients at the $t$-th epoch to form a global style pool, denoted by $\bm{\mu}_{\text{style}}^{(t)} = [\mu_1^{(t)}, \mu_2^{(t)}, \ldots, \mu_m^{(t)}], \bm{\sigma}_{\text{style}}^{(t)} = [\sigma_1^{(t)}, \sigma_2^{(t)}, \ldots, \sigma_m^{(t)}],$ where $M$ is the total number of clients. These global style statistics are shared among clients, allowing for the collaborative alignment of stylistic patterns and the enhancement of local data into more comprehensive feature representations. In the $(t+1)$-th epoch, each client $m$ performs style transfer \citep{huang2017adain} using the shared statistics from the global style pool, which is particularly well suited for our purposes due to its efficiency and compact representation. The transfer is mathematically expressed as:

\begin{equation}
\bm{x}_{m, style}^{(t+1)} = \left( \frac{\bm{x}_m^{(t+1)} - \mu_m^{(t+1)}}{\sigma_m^{(t+1)}} \right) \bm{\sigma}_{\text{style}}^{(t)}  + \bm{\mu}_{\text{style}}^{(t)},
\end{equation}

where $\mu_m^{(t)}$ and $\sigma_m^{(t)}$ are the mean and standard deviation of client $m$’s images at the $t$-th epoch. This exchange of style information among clients enhances the model’s generalization ability in heterogeneous FL environments by introducing variability in image-level representations.

\begin{figure*}[htp]
\centering
\includegraphics[width=0.5\textwidth]{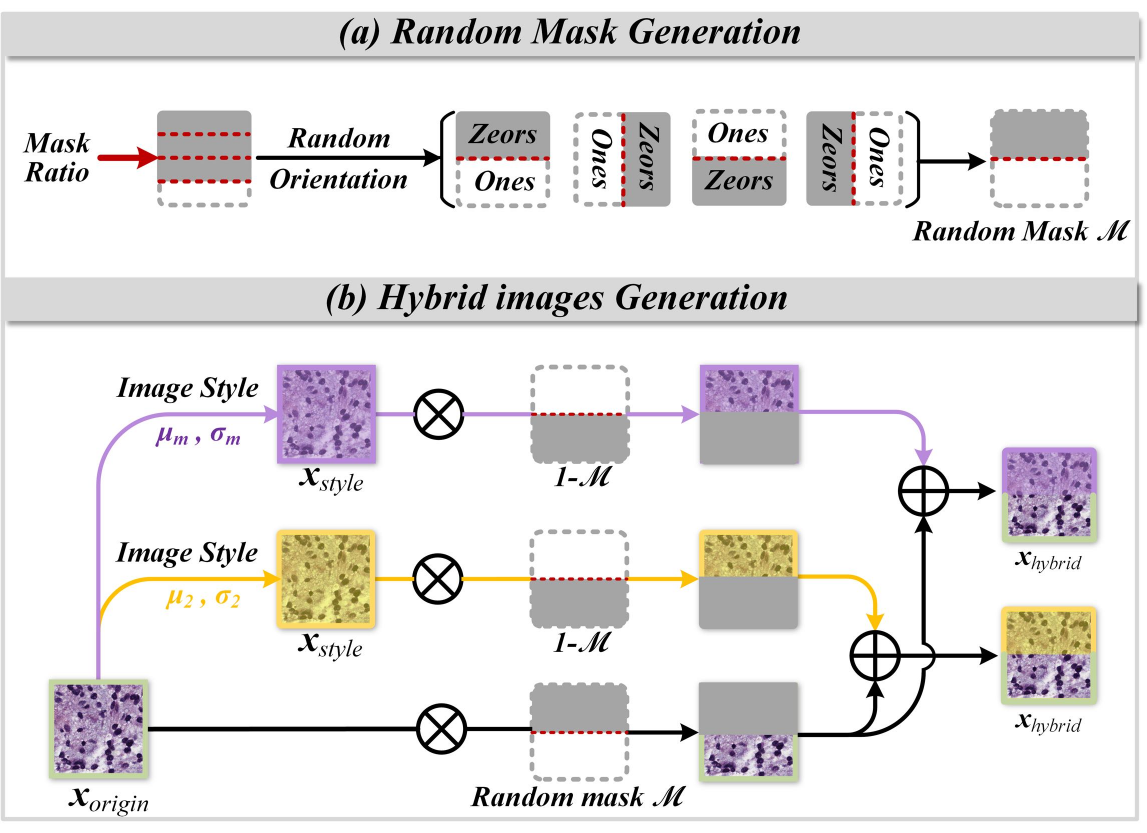}
\caption{ The flowchart illustrates the generation process of random masks and hybrid images.}
\label{fig_mask}
\end{figure*}

In the collaborative style enhancement module, we also introduce a random mask to balance the fidelity of the original image $\bm{x}_{origin}$ with the integration of stylistic variations $\bm{x}_{style} \in \{\bm{x}_{m,j} \}_{j=1}^M$. { The random mask $\mathcal{M}$ is generated as a binary matrix with equal halves of zeros and ones, followed by random rotations of 0°, 90°, 180°, and 270°, including horizontal and vertical flips, as illustrated in Fig.~\ref{fig_mask} (a). This ensures a balanced hybrid ratio between original and style-transferred images to generate hybrid images in Fig.~\ref{fig_mask} (b).} Specifically, the hybrid image $\bm{x}_{hybrid}$ is achieved using the following formula:
\begin{equation} 
\bm{x}_{hybrid} = \mathcal{M} \otimes \bm{x}_{m} + (1 - \mathcal{M}) \otimes \bm{x}_{style}
\end{equation}
where $\otimes$ denotes the pixel-wise multiplication operator. This masking allows the model to retain the original semantic features while integrating style information from different clients. The combined representation captures both the inherent content and diverse stylistic features, enabling the model to improve segmentation accuracy across heterogeneous datasets.

\subsection{Adaptive Feature Alignment}
At the feature level, we propose an adaptive feature alignment module to ensure implicit alignment within the representation space by effectively integrating local feature details with global contextual information. Feature-level representations capture high-level semantic information essential for pathology image segmentation, enabling a better understanding of the contextual relationships between different structures within the image \citep{song2022global}. Each client’s feature representations retain the unique characteristics of their local data while being aligned with global mean style statistics, promoting semantic consistency across clients. 

Specifically, we calculate the average mean ${\mu}^{z(t)}$ and standard deviation ${\sigma}^{z(t)}$ of the deepest level feature maps $Z_m$ at the $t$-th training round, which represents the global distribution of high-level features across all centers. The mean and standard deviation are computed as follows:
\begin{equation}
{\mu}^{z(t)} = \frac{1}{MN} \sum_{m=1}^{M} \sum_{i=1}^{N} Z_m^{(t)}(i),
\label{eq:feature_mu}
\end{equation}
\begin{equation}
{\sigma}^{z(t)} =\frac{1}{M} \sum_{m=1}^{M}  \sqrt{\frac{1}{N} \sum_{i=1}^{N} (Z_m^{(t)}(i) - \mu_m^{z(t)})^2},
\label{eq:feature_std}
\end{equation}
where $N$ is the total number of feature elements in $Z_m^{(t)}$. The client $m$ incorporates the statistics from the global feature style, thereby adapting its own feature representations of mean $\mu_m^{z(t)}$ and standard deviation $\sigma_m^{z(t)}$ and aligning high-level semantic representations by infusing local feature details with global contextual insights. The core formula for feature alignment is given by:
\begin{equation}
Z_{m, style}^{(t+1)} ={\sigma}^{z(t)} \left( \frac{Z_m^{(t+1)} - \mu_m^{z(t+1)}}{\sigma_m^{z(t+1)}} \right) + {\mu}^{z(t)},
\end{equation}
where $Z_m^{(t)}$ and $Z_m^{(t+1)}$ represent the feature maps before and after the style alignment process, respectively. This implicit alignment harmonizes client-specific feature distributions within a shared semantic space, facilitating a cohesive understanding of structural information. By seamlessly integrating local features with global style information, the adaptive feature alignment module improves the model's ability to capture fine-grained boundary details and understand contextual relationships in multi-center heterogeneous data.

Overall, the collaborative style enhancement module enhances texture detail information at the image level, broadening the diversity of unseen visual patterns across different centers. Meanwhile, the adaptive feature alignment module aligns semantic representations at the feature level, promoting consistency and reducing representation bias. Together, these two modules effectively mitigate heterogeneity and enhance overall model performance in segmentation tasks.

\begin{table*}[t]
\caption{Detailed information of four sets of heterogeneous pathology image datasets involved in this study.}
\centering 
\setlength{\tabcolsep}{3.2pt} 
\label{tab: table_dataset}
\renewcommand{\arraystretch}{1.2} 
\resizebox{\textwidth}{!}{%
\begin{tabular}{|c|c|c|c|c|c|c|c|} 
\hline
\hline
\multicolumn{8}{|c|}{\rule{0pt}{1.5em}\textbf{Cross-source datasets}} \\ \hline \hline
\multicolumn{1}{|c|}{\textbf{Client}} & \multicolumn{1}{c|}{\textbf{Dataset}} & \multicolumn{1}{c|}{\textbf{Organ}} & \multicolumn{1}{c|}{\textbf{Origin}} & \multicolumn{1}{c|}{\textbf{Scanner}} & \multicolumn{1}{c|}{\textbf{Magnification}} & \multicolumn{1}{c|}{\textbf{Size}} & \textbf{Number} \\ \hline
\multicolumn{1}{|c|}{C1} & \multicolumn{1}{c|}{CoNSep} & \multicolumn{1}{c|}{Colon} & \multicolumn{1}{c|}{UHCW} & \multicolumn{1}{c|}{Omnyx VL120   scanner} & \multicolumn{1}{c|}{40 ×} & \multicolumn{1}{c|}{1000 × 1000} & 41 \\ \hline
\multicolumn{1}{|c|}{C2} & \multicolumn{1}{c|}{CPM17} & \multicolumn{1}{c|}{Multiple} & \multicolumn{1}{c|}{TCGA} & \multicolumn{1}{c|}{/} & \multicolumn{1}{c|}{20 × and 40 ×} & \multicolumn{1}{c|}{500 × 500 and 600 ×   600} & 32 \\ \hline
\multicolumn{1}{|c|}{C3} & \multicolumn{1}{c|}{CRAG} & \multicolumn{1}{c|}{Colon} & \multicolumn{1}{c|}{UHCW} & \multicolumn{1}{c|}{Omnyx VL120 scanner} & \multicolumn{1}{c|}{20 ×} & \multicolumn{1}{c|}{1512×1516} & 213 \\ \hline
\multicolumn{1}{|c|}{C4} & \multicolumn{1}{c|}{CryoNuSeg} & \multicolumn{1}{c|}{Multiple} & \multicolumn{1}{c|}{TCGA} & \multicolumn{1}{c|}{/} & \multicolumn{1}{c|}{40 ×} & \multicolumn{1}{c|}{512 × 512} & 30 \\ \hline
\multicolumn{1}{|c|}{C5} & \multicolumn{1}{c|}{Glas} & \multicolumn{1}{c|}{Colon} & \multicolumn{1}{c|}{UHCW} & \multicolumn{1}{c|}{Zeiss MIRAX MIDI scanner} & \multicolumn{1}{c|}{20 ×} & \multicolumn{1}{c|}{775 × 522} & 165 \\ \hline
\multicolumn{1}{|c|}{C6} & \multicolumn{1}{c|}{KUMAR} & \multicolumn{1}{c|}{Multiple} & \multicolumn{1}{c|}{TCGA} & \multicolumn{1}{c|}{/} & \multicolumn{1}{c|}{40 ×} & \multicolumn{1}{c|}{1000 × 1000} & 30 \\ \hline
\multicolumn{1}{|c|}{C7} & \multicolumn{1}{c|}{TNBC} & \multicolumn{1}{c|}{Breast} & \multicolumn{1}{c|}{Curie Institute} & \multicolumn{1}{c|}{Philips ultra fast scanner} & \multicolumn{1}{c|}{40 ×} & \multicolumn{1}{c|}{512 × 512} & 50 \\ 
\hline
\hline
\multicolumn{8}{|c|}{\rule{0pt}{1.5em}\textbf{Cross-modality datasets}} \\ \hline \hline
\multicolumn{1}{|c|}{\textbf{Client}} & \multicolumn{1}{c|}{\textbf{Dataset}} & \multicolumn{1}{c|}{\textbf{Organ}} & \multicolumn{1}{c|}{\textbf{Modality}} & \multicolumn{1}{c|}{\textbf{Scanner}} & \multicolumn{1}{c|}{\textbf{Magnification}} & \multicolumn{1}{c|}{\textbf{Size}} & \textbf{Number} \\ \hline
\multicolumn{1}{|c|}{C1} & \multicolumn{1}{c|}{\multirow{3}{*}{ClusterSeg}} & \multicolumn{1}{c|}{Thyroid} & \multicolumn{1}{c|}{Cytology} & \multicolumn{1}{c|}{3CCD linear camera} & \multicolumn{1}{c|}{20 ×} & \multicolumn{1}{c|}{224 × 224} & 487 \\ \cline{1-1} \cline{3-8} 
\multicolumn{1}{|c|}{C2} & \multicolumn{1}{c|}{} & \multicolumn{1}{c|}{Embryos and archaea} & \multicolumn{1}{c|}{Fluorescence} & \multicolumn{1}{c|}{Zeiss confocal   microscope} & \multicolumn{1}{c|}{63 ×} & \multicolumn{1}{c|}{512 × 512} & 524 \\ \cline{1-1} \cline{3-8} 
\multicolumn{1}{|c|}{C3} & \multicolumn{1}{c|}{} & \multicolumn{1}{c|}{Colorectal} & \multicolumn{1}{c|}{Histology} & \multicolumn{1}{c|}{/} & \multicolumn{1}{c|}{40 ×} & \multicolumn{1}{c|}{512 × 512} & 462 \\ \hline
\hline
\multicolumn{8}{|c|}{\rule{0pt}{1.5em}\textbf{Cross-organ datasets}} \\ \hline \hline
\multicolumn{1}{|c|}{\textbf{Client}} & \multicolumn{1}{c|}{\textbf{Dataset}} & \multicolumn{1}{c|}{\textbf{Organ}} & \multicolumn{1}{c|}{\textbf{Modality}} & \multicolumn{1}{c|}{\textbf{Scanner}} & \multicolumn{1}{c|}{\textbf{Magnification}} & \multicolumn{1}{c|}{\textbf{Size}} & \textbf{Number} \\ \hline
\multicolumn{1}{|c|}{C1} & \multicolumn{1}{c|}{\multirow{3}{*}{COSAS2024 Task 1}} & \multicolumn{1}{c|}{Colorectum} & \multicolumn{1}{c|}{\multirow{3}{*}{Histology}} & \multicolumn{1}{c|}{TEKSQRAY   SQS-600P} & \multicolumn{1}{c|}{/} & \multicolumn{1}{c|}{512×512} & 198 \\ \cline{1-1} \cline{3-3} \cline{5-8} 
\multicolumn{1}{|c|}{C2} & \multicolumn{1}{c|}{} & \multicolumn{1}{c|}{Pancreas} & \multicolumn{1}{c|}{} & \multicolumn{1}{c|}{TEKSQRAY   SQS-600P} & \multicolumn{1}{c|}{/} & \multicolumn{1}{c|}{512×512} & 138 \\ \cline{1-1} \cline{3-3} \cline{5-8} 
\multicolumn{1}{|c|}{C3} & \multicolumn{1}{c|}{} & \multicolumn{1}{c|}{Stomach} & \multicolumn{1}{c|}{} & \multicolumn{1}{c|}{TEKSQRAY   SQS-600P} & \multicolumn{1}{c|}{/} & \multicolumn{1}{c|}{512×512} & 141 \\ \hline
\hline
\multicolumn{8}{|c|}{\rule{0pt}{1.5em}\textbf{Cross-scanner datasets}} \\ \hline \hline
\multicolumn{1}{|c|}{\textbf{Client}} & \multicolumn{1}{c|}{\textbf{Dataset}} & \multicolumn{1}{c|}{\textbf{Organ}} & \multicolumn{1}{c|}{\textbf{Modality}} & \multicolumn{1}{c|}{\textbf{Scanner}} & \multicolumn{1}{c|}{\textbf{Magnification}} & \multicolumn{1}{c|}{\textbf{Size}} & \textbf{Number} \\ \hline
\multicolumn{1}{|c|}{C1} & \multicolumn{1}{c|}{\multirow{3}{*}{COSAS2024 Task 2}} & \multicolumn{1}{c|}{Breast} & \multicolumn{1}{c|}{\multirow{3}{*}{Histology}} & \multicolumn{1}{c|}{3d-1000} & \multicolumn{1}{c|}{/} & \multicolumn{1}{c|}{512×512} & 203 \\ \cline{1-1} \cline{3-3} \cline{5-8} 
\multicolumn{1}{|c|}{C2} & \multicolumn{1}{c|}{} & \multicolumn{1}{c|}{Breast} & \multicolumn{1}{c|}{} & \multicolumn{1}{c|}{Kfbio-400} & \multicolumn{1}{c|}{/} & \multicolumn{1}{c|}{512×512} & 193 \\ \cline{1-1} \cline{3-3} \cline{5-8} 
\multicolumn{1}{|c|}{C3} & \multicolumn{1}{c|}{} & \multicolumn{1}{c|}{Breast} & \multicolumn{1}{c|}{} & \multicolumn{1}{c|}{Teksqray-600p} & \multicolumn{1}{c|}{/} & \multicolumn{1}{c|}{512×512} & 205 \\ \hline
\end{tabular}}

\end{table*}

\subsection{Stratified Similarity Aggregation}
We propose a stratified similarity aggregation strategy using Gaussian-distributed synthetic data to evaluate the similarity between layers of the network for better preservation of critical information while minimizing the impact of heterogeneity. For each client $m$, the mean $\overline{\mu}_m$ and standard deviation $\overline{\sigma}_m$ characterize the distribution of image intensities computed over its entire local dataset. During the $t_{th}$ communication round, Gaussian synthetic data $g_m \sim\mathcal{N} (\overline{\mu}_m,\overline{\sigma}_m)$ is randomly generated and then uploaded to the server to facilitate similarity evaluation among clients. The entire client network is stratified into $L$ layers, each exhibiting distinct feature representation capabilities. By employing this stratified similarity aggregation strategy, we achieve a more refined and flexible aggregation process that hierarchically aligns and dynamically integrates weights from both shallow and deep layers, enhancing size-specific representations amid heterogeneous data.

Let the feature output of client $m$ at layer $l$ be denoted as $\phi_m^l(g_m)$, obtained by passing the Gaussian synthetic data through the client's model up to layer $l$. For client $m$ and $j$, the cosine similarity between layer $l$ feature outputs is defined as $\text{CosSim}_{m,j}^{l} = \frac{\phi_m^l(g_m) \cdot \phi_j^l(g_j)}{| \phi_m^l(g_m) | | \phi_j^l(g_j) |}$. Subsequently, the cosine similarity between each pair of layers is scaled by the total sum of all pairwise cosine similarities, denoted as $s_{m,j}^l$. This normalization ensures that the values are relative to the overall similarity distribution across all client pairs. Next, the weights of the server model are aggregated layer by layer based on the similarity evaluation interaction between clients' models. During the $t$ communication round, the server model $\bm{w}^{t,l}$ aggregates the weights from layer $l$ of the $t-1$ round of client models with corresponding interaction weights as follows:

\begin{equation}
	{\bm{w}_{server}^{l(t+1)}}=\frac{1}{2M} \sum_{m=1}^M \sum_{j=1, j\neq m}^M [{\bm{w}_m^{(t)}}+\sum_{j\neq k} {s^l_{m,j} \cdot {\bm{w}_j^{(t)}} }].  
\end{equation} 

This hierarchical aggregation process culminates in the updated server model $\bm{w}_{server}^{t}$, which is subsequently distributed to each client for the next training epoch, expressed as $\bm{w}_m^{t+1} \leftarrow{} \bm{w}_{server}^{t}$. The adoption of a stratified similarity aggregation strategy allows for the allocation of greater weight to layers that exhibit higher similarity during the model aggregation process, while assigning lower weight to layers that demonstrate lower similarity. By emphasizing layers with high similarity, the model enhances its ability to capture shared features, thereby improving overall performance and accuracy, while simultaneously mitigating the influence of low-similarity layers, which reduces noise and irrelevant information during aggregation. This improves the interactive flexibility of the server model, ensuring more effective model alignment. As a result, this strategy supports enhanced generalization capabilities and performance across diverse and heterogeneous data environments.

\section{Results} 

\subsection{Datasets}
In our experiments, we validated the algorithm on four sets of multi-center datasets, each reflecting different dimensions of heterogeneity: cross-source, cross-modality, cross-organ, and cross-scanner. These four sets were selected to simulate the various heterogeneity challenges commonly encountered in real-world pathology clinical environments, where data sources (cross-source), imaging techniques (cross-modality), anatomical regions (cross-organ), and scanning devices (cross-scanner) vary significantly. a) \textbf{Cross-source datasets}: seven independent datasets of hematoxylin and eosin (HE) stained pathology images from different sources: CoNSeP \citep{graham2019hover}, CPM-17 \citep{vu2019cpm17}, CRAG \citep{graham2019crag}, CryoNuSeg \citep{mahbod2021cryonuseg}, Glas \citep{SIRINUKUNWATTANA2017489glas}, Kumar \citep{kumar2017dataset}, and TNBC \citep{naylor2018tnbc}. The cross-source datasets encompass different origins, disease types, organs, scanning devices, and resolutions, representing the most complex and realistic scenarios. b) \textbf{Cross-modality datasets}: three datasets from ClusterSeg \citep{ke2023clusterseg}  with characteristic severely clustered nuclei across three image modalities: cytology, fluorescence, and histology. c) \textbf{Cross-organ datasets}: three datasets from COSAS2024 Task 1 \citep{grandchallenge2024cosas} with the same adenocarcinoma across three organs: colorectum, pancreas, and stomach. d) \textbf{Cross-scanner datasets}: three datasets from COSAS2024 Task 2 \citep{grandchallenge2024cosas} with the same organ scanned by three distinct WSI scanners. Samples from COSAS2024 are recropped into $512 \times 512$ patches, excluding empty background patches. { e) \textbf{Unified datasets}: a combination of all 16 datasets from Table 1.} All datasets were then split into training and test sets following their original dataset partition protocol or with a ratio of $8:2$, maintaining this splitting strategy consistently across all experiments to ensure fair comparison. The detailed information for the four sets of datasets is summarized in Table \ref{tab: table_dataset} and each dataset is allocated to a single client.
\begin{table*}[t]
\centering
\caption{Cross-source results across different models and centers. The bold indicates the best value.}
\label{tabel: result_cross_source}
\renewcommand{\arraystretch}{1.8} 
\resizebox{\textwidth}{!}{%
\begin{tabular}{rllllllllll||llllllllll}
\toprule
        & \multicolumn{10}{c||}{\textbf{Dice (\%) $\uparrow$}}   & \multicolumn{10}{c}{\textbf{ASSD \textit{(pix.)} $\downarrow$}}    \\ 
\cmidrule(lr){2-11} \cmidrule(lr){12-21}
        & C1           & C2           & C3           & C4           & C5           & C6           & C7           & Average   & \textcolor{black}{p-value} & \textcolor{black}{95\% CI}          & C1          & C2          & C3            & C4          & C5            & C6          & C7          & Average   & \textcolor{black}{p-value} & \textcolor{black}{95\% CI} \\ 
\midrule
FedAvg \citep{McMahan2017fedavg} & 80.48\textsubscript{±3.82} & 87.01\textsubscript{±3.25} & 84.09\textsubscript{±8.15} & 81.81\textsubscript{±2.26} & 86.97\textsubscript{±8.21} & 79.99\textsubscript{±3.29} & 78.01\textsubscript{±5.96} & 82.62\textsubscript{±4.99} & \textcolor{black}{0.0013} & \textcolor{black}{[80.21, 84.97]} & 2.89\textsubscript{±0.74} & 2.33\textsubscript{±0.93} & 46.23\textsubscript{±42.14} & 2.64\textsubscript{±0.22} &  19.27\textsubscript{±18.41} & 3.42\textsubscript{±0.74} & 6.01\textsubscript{±4.80} & 11.83\textsubscript{±9.71} & \textcolor{black}{0.0436} & \textcolor{black}{[3.26, 17.50]} \\
FedBN \citep{li2021fedbn} & 81.46\textsubscript{±3.59} & 86.94\textsubscript{±3.94} & 82.24\textsubscript{±5.56} & 80.92\textsubscript{±2.03} & 87.63\textsubscript{±6.80} & 79.78\textsubscript{±4.13} & \textbf{79.93}\textsubscript{±3.87} & 82.70\textsubscript{±4.28} & \textcolor{black}{0.0524} & \textcolor{black}{[80.55, 84.79]} & 2.95\textsubscript{±0.85} & 2.23\textsubscript{±0.71} & 39.18\textsubscript{±23.39} & 3.17\textsubscript{±0.47} & 17.89\textsubscript{±11.88} & 3.52\textsubscript{±0.82} & \textbf{5.61}\textsubscript{±4.43} & 10.65\textsubscript{±6.08} & \textcolor{black}{0.0985} & \textcolor{black}{[3.54, 14.45]} \\
FedProx \citep{li2020fedProx} & 79.82\textsubscript{±4.09} & 86.79\textsubscript{±3.58} & 83.37\textsubscript{±7.56} & 80.89\textsubscript{±2.66} & 87.28\textsubscript{±7.93} & 79.58\textsubscript{±3.90} & 77.73\textsubscript{±5.53} & 82.21\textsubscript{±5.04} & \textcolor{black}{0.0017} & \textcolor{black}{[79.65, 84.62]} & 3.32\textsubscript{±1.51} & 2.41\textsubscript{±1.02} & 41.38\textsubscript{±30.96} & 2.87\textsubscript{±0.16} & 15.43\textsubscript{±10.66} & 3.39\textsubscript{±0.78} & 6.06\textsubscript{±4.02} & 10.69\textsubscript{±7.01} & \textcolor{black}{0.0224} & \textcolor{black}{[3.13, 14.73]} \\
HarmoFL \citep{jiang2022harmofl} & 79.69\textsubscript{±3.79} & 86.65\textsubscript{±3.39} & 80.33\textsubscript{±7.67} & 81.12\textsubscript{±4.15} & 86.54\textsubscript{±7.60} & 80.88\textsubscript{±2.35} & 78.38\textsubscript{±5.75} & 81.94\textsubscript{±4.96} & \textcolor{black}{0.0171} & \textcolor{black}{[79.86, 84.09]} & 4.49\textsubscript{±4.63} & 2.42\textsubscript{±0.96} & 44.14\textsubscript{±34.98} & 2.55\textsubscript{±0.55} & 17.37\textsubscript{±13.03} & 3.21\textsubscript{±0.61} & 6.55\textsubscript{±5.09} & 11.53\textsubscript{±8.55} & \textcolor{black}{0.0216} & \textcolor{black}{[3.24, 16.68]} \\
HistoFL \citep{lu2022histofl} & 80.32\textsubscript{±3.26} & 87.01\textsubscript{±3.44} & 83.68\textsubscript{±7.34} & 81.60\textsubscript{±3.04} & 87.69\textsubscript{±7.38} & 80.37\textsubscript{±3.28} & 78.37\textsubscript{±5.95} & 82.72\textsubscript{±4.81} & \textcolor{black}{0.0049} & \textcolor{black}{[80.13, 85.21]} & 3.23\textsubscript{±1.27} & 2.38\textsubscript{±0.96} & 44.19\textsubscript{±33.38} & 2.70\textsubscript{±0.17} & 16.96\textsubscript{±13.14} & 3.26\textsubscript{±0.69} & 6.80\textsubscript{±5.23} & 11.36\textsubscript{±7.83} & \textcolor{black}{0.0290} & \textcolor{black}{[3.29, 15.84]} \\
FedFA \citep{zhou2023fedfa} & 80.11\textsubscript{±3.75} & 86.80\textsubscript{±3.23} & 83.99\textsubscript{±6.20} & 81.20\textsubscript{±3.39} & 86.21\textsubscript{±7.92} & 80.32\textsubscript{±3.19} & 78.07\textsubscript{±6.70} & 82.39\textsubscript{±4.91} & \textcolor{black}{0.0027} & \textcolor{black}{[80.22, 84.59]} & 3.34\textsubscript{±1.16} & 2.30\textsubscript{±0.82} & 42.21\textsubscript{±33.12} & 2.73\textsubscript{±0.17} & 18.79\textsubscript{±18.52} & 3.23\textsubscript{±0.67} & 8.06\textsubscript{±8.02} & 11.52\textsubscript{±8.93} & \textcolor{black}{0.0114} & \textcolor{black}{[3.60, 15.67]} \\
FedHEAL \citep{chen2024fair} & 80.48\textsubscript{±3.58} & 86.89\textsubscript{±3.45} & 
83.86\textsubscript{±8.50} & 
81.75\textsubscript{±3.43} & 
87.33\textsubscript{±7.27} & 
80.19\textsubscript{±3.87} & 
77.64\textsubscript{±5.87} & 
82.59\textsubscript{±5.14} & \textcolor{black}{0.0011} & \textcolor{black}{[80.25, 84.98]} &  
2.95\textsubscript{±0.75} &
2.29\textsubscript{±0.84} & 
38.10\textsubscript{±29.63} & 
2.65\textsubscript{±0.15} & 
17.45\textsubscript{±14.05} & 
3.25\textsubscript{±0.76} & 
8.04\textsubscript{±7.89} & 
10.68\textsubscript{±7.73} & \textcolor{black}{0.0079} & \textcolor{black}{[3.23, 14.72]} \\
\textbf{PathFL (Ours)}  & \textbf{82.06}\textsubscript{±3.57} & \textbf{87.46}\textsubscript{±3.24} & \textbf{86.12}\textsubscript{±5.02} & \textbf{84.72}\textsubscript{±3.44} & \textbf{88.96}\textsubscript{±8.27} & \textbf{81.44}\textsubscript{±2.75} & 79.17\textsubscript{±4.70} & \textbf{84.27}\textsubscript{±4.43} & \textcolor{black}{-} & \textcolor{black}{[81.72, 86.73]} & \textbf{2.76}\textsubscript{±0.73} & \textbf{2.16}\textsubscript{±0.77} & \textbf{36.47}\textsubscript{±26.28} & \textbf{2.14}\textsubscript{±0.46} & \textbf{13.96}\textsubscript{±10.80} & \textbf{2.99}\textsubscript{±0.53} & 5.94\textsubscript{±4.92} & \textbf{9.49}\textsubscript{±6.36} & \textcolor{black}{-} & \textcolor{black}{[2.53, 13.20]} \\
\bottomrule  
\end{tabular}%
}
\end{table*}

\begin{table*}[t]
\centering
\caption{Cross-modality results across different models and centers. The bold indicates the best value.}
\setlength{\tabcolsep}{4.5pt} 
\label{tabel: result_cross_modality} 
\renewcommand{\arraystretch}{1.2} 
\resizebox{\textwidth}{!}{%
\begin{tabular}{rllllll||llllll}
\toprule
        & \multicolumn{6}{c||}{\textbf{Dice (\%) $\uparrow$}}                    & \multicolumn{6}{c}{\textbf{ASSD \textit{(pix.)} $\downarrow$}}  \\
\cmidrule(lr){2-7} \cmidrule(lr){8-13}        
        & C1            & C2           & C3           & Average      & \textcolor{black}{p-value} & \textcolor{black}{95\% CI} & C1           & C2            & C3           & Average      & \textcolor{black}{p-value} & \textcolor{black}{95\% CI} \\
\midrule        
FedAvg \citep{McMahan2017fedavg} & 83.41\textsubscript{±9.85}  & 91.46\textsubscript{±4.64} & 73.53\textsubscript{±6.71} & 82.80\textsubscript{±7.07} & \textcolor{black}{0.0240} & \textcolor{black}{[76.00, 89.45]} & 6.22\textsubscript{±8.70}  & 6.72\textsubscript{±4.42}  & 9.36\textsubscript{±4.66} & 7.43\textsubscript{±5.93} & \textcolor{black}{0.0271} & \textcolor{black}{[6.47, 8.70]}  \\
FedBN \citep{li2021fedbn} & 87.07\textsubscript{±5.24}  & 90.45\textsubscript{±4.66} & 77.38\textsubscript{±5.29} & 84.97\textsubscript{±5.06} & \textcolor{black}{0.0069} & \textcolor{black}{[79.80, 89.08]} & 4.47\textsubscript{±3.69}  & 12.49\textsubscript{±11.89} & 7.69\textsubscript{±3.11} & 8.21\textsubscript{±6.23} & \textcolor{black}{0.0755} & \textcolor{black}{[5.40, 11.29]} \\
FedProx \citep{li2020fedProx} & 73.20\textsubscript{±11.98} & 87.82\textsubscript{±7.55} & 63.00\textsubscript{±7.67} & 74.67\textsubscript{±9.06} & \textcolor{black}{0.0173} & \textcolor{black}{[65.91, 84.53]} & 13.07\textsubscript{±7.73} & 17.44\textsubscript{±19.06} & 15.00\textsubscript{±6.08} & 15.17\textsubscript{±10.96} & \textcolor{black}{0.0002} & \textcolor{black}{[13.60, 16.87]} \\
HarmoFL \citep{jiang2022harmofl} & 81.80\textsubscript{±6.85}  & 89.04\textsubscript{±5.75} & 69.55\textsubscript{±6.88} & 80.13\textsubscript{±6.49} & \textcolor{black}{0.0157} & \textcolor{black}{[72.61, 85.45]} & 8.55\textsubscript{±8.79}  & 11.12\textsubscript{±6.74}  & 11.38\textsubscript{±4.97} & 10.35\textsubscript{±6.83} & \textcolor{black}{0.0001} & \textcolor{black}{[9.19, 11.25]} \\
HistoFL \citep{lu2022histofl} & 83.01\textsubscript{±9.33}  & 92.74\textsubscript{±3.68} & 75.59\textsubscript{±6.91} & 83.78\textsubscript{±6.64} & \textcolor{black}{0.0363} & \textcolor{black}{[77.64, 90.31]} & 7.15\textsubscript{±8.91}  & 7.59\textsubscript{±11.42}  & 8.48\textsubscript{±4.50}  & 7.74\textsubscript{±8.28} & \textcolor{black}{0.0052} & \textcolor{black}{[7.30, 8.26]} \\
FedFA \citep{zhou2023fedfa} & 83.32\textsubscript{±7.31}  & 90.69\textsubscript{±4.53} & 73.49\textsubscript{±6.15} & 82.50\textsubscript{±6.00} & \textcolor{black}{0.0156} & \textcolor{black}{[75.95, 88.65]} & 7.58\textsubscript{±8.93}  & 10.40\textsubscript{±11.49} & 9.75\textsubscript{±4.71}  & 9.24\textsubscript{±8.38} & \textcolor{black}{3.798E-05} & \textcolor{black}{[8.12, 10.07]} \\
FedHEAL \citep{chen2024fair} & 83.22\textsubscript{±9.49}  & 91.68\textsubscript{±4.30} & 
74.53\textsubscript{±6.59} & 
83.14\textsubscript{±6.79} & \textcolor{black}{0.0220} & \textcolor{black}{[76.70, 89.55]} & 
5.51\textsubscript{±6.83} & 
9.91\textsubscript{±13.10} & 
8.90\textsubscript{±4.39} & 
8.11\textsubscript{±8.11} & \textcolor{black}{0.0056} & \textcolor{black}{[6.36, 9.46]} \\
\textbf{PathFL (Ours)}  & \textbf{89.93}\textsubscript{±3.49}  & \textbf{93.88}\textsubscript{±3.10} & \textbf{83.00}\textsubscript{±5.30} & \textbf{88.93}\textsubscript{±3.96} & \textcolor{black}{-} & \textcolor{black}{[84.48, 91.92]} & \textbf{3.54}\textsubscript{±4.62}  & \textbf{5.81}\textsubscript{±6.69}   & \textbf{5.44}\textsubscript{±3.60}  & \textbf{4.93}\textsubscript{±4.97} & \textcolor{black}{-} & \textcolor{black}{[4.01, 5.63]} \\

\bottomrule  
\end{tabular}%
}
\end{table*}

\subsection{Implementation Details}
In this work, we employ the U-Net \citep{ronneberger2015u} as the basic client network, which is trained over 300 epochs with 5 local update epochs per communication round in all experiments involving cross-source, cross-organ, and cross-scanner data settings. For cross-modality datasets, training sets to 500 epochs, maintaining a 5-epoch communication round structure. To ensure fairness, all comparative experiments are initialized identically, with the same communication round configurations across experiments. For training, we consistently use a batch size of 4 and the Adam optimizer, configured with a learning rate of $1e^{-4}$ and momentum parameters set to 0.9 and 0.95. To ensure fairness in model training, we consistently apply Cross Entropy Loss as the unified loss function to evaluate the discrepancy between predicted segmentation results and ground truth labels. Our experiments are executed on a cluster of four 40 GB NVIDIA A6000 GPUs. All implementations are completed using PyTorch.

To comprehensively evaluate segmentation performance, we employ two complementary metrics: the Dice similarity coefficient (Dice) and the Average Symmetric Surface Distance (ASSD) \citep{yeghiazaryan2018family}. The Dice coefficient, widely used in medical image segmentation, quantifies the overlap between predictions and labels. By balancing sensitivity and precision, it provides a robust measure of overall segmentation similarity. ASSD focuses on boundary alignment, capturing the average discrepancy between predicted and ground truth surfaces. This makes it particularly valuable for assessing the precision of segmentation boundaries, especially in high-resolution medical images where accurate delineation is critical. Together, these metrics offer a holistic segmentation evaluation. { We used paired t-tests and 95\% confidence intervals to evaluate statistical significance, with p-value $<$ 0.05 indicating significant differences between methods.}

\subsection{Comparison with the State-of-the-arts}
For a comprehensive evaluation, we benchmark our approach against state-of-the-art (SOTA) FL methods designed to address heterogeneity. The comparison includes classic methods like FedAvg, pathology-specific approaches such as HistoFL, input-based techniques like HarmoFL, and feature-level strategies including FedBN and FedFA. Additionally, we evaluate aggregation-focused methods such as FedProx, HistoFL, and FedHEAL to assess their effectiveness in managing task heterogeneity within federated learning frameworks.

\subsubsection{Performance on cross-source datasets}
Our proposed model PathFL achieves the best performance on the cross-source datasets, attaining an average Dice of 84.27\% and an average ASSD of 9.49 with notable improvements of 1.55\% in the Dice and 1.65 in ASSD compared to the baseline (FedAvg). The quantitative segmentation results of Dice and ASSD on the cross-source datasets for seven different centers (C1 to C7) are summarized in Table \ref{tabel: result_cross_source}. The cross-source dataset represents the most challenging clinical scenario, characterized by the largest number of centers, the highest degree of data imbalance, and significant inter-center heterogeneity. Importantly, our method demonstrates consistent performance across all centers without sacrificing the accuracy of any specific center for the benefit of another. This balanced performance validates that our approach can adapt effectively to heterogeneous clients, providing a tailored yet generalizable solution across diverse data distributions without compromising the segmentation quality of any center. Furthermore, this performance is particularly notable in most centers, especially in C3 and C5, with Dice of 86.12\% and 88.96\%, respectively. Both C3 and C5 correspond to datasets derived from colon organs, highlighting the capability of our method to not only mitigate the interference caused by heterogeneity but also to effectively identify and leverage clients with representational similarities for more efficient aggregation. In conclusion, PathFL effectively addresses inter-center heterogeneity and data imbalance across cross-source datasets. By leveraging representational similarities, PathFL achieves balanced performance and demonstrates promising generalizability in diverse clinical settings.

\begin{table*}[!t]
\centering
\caption{Cross-organ results across different models and centers. The bold indicates the best value.}
\setlength{\tabcolsep}{3.5pt} 
\label{tabel: result_cross_organ} 
\renewcommand{\arraystretch}{1.2} 
\resizebox{\textwidth}{!}{%
\begin{tabular}{rllllll||llllll}
\toprule
        & \multicolumn{6}{c||}{\textbf{Dice (\%) $\uparrow$}}                    & \multicolumn{6}{c}{\textbf{ASSD \textit{(pix.)} $\downarrow$}}  \\
\cmidrule(lr){2-7} \cmidrule(lr){8-13}        
        & C1            & C2           & C3           & Average      & \textcolor{black}{p-value} & \textcolor{black}{95\% CI} & C1           & C2            & C3           & Average       & \textcolor{black}{p-value} & \textcolor{black}{95\% CI} \\
\midrule        
FedAvg \citep{McMahan2017fedavg} & 79.08\textsubscript{±16.72} & 72.79\textsubscript{±20.83} & 73.47\textsubscript{±22.67} & 75.11\textsubscript{±20.07} & \textcolor{black}{0.0264} & \textcolor{black}{[72.79, 79.08]} & 65.39\textsubscript{±56.70} & 87.61\textsubscript{±69.36} & 91.98\textsubscript{±69.44} & 81.66\textsubscript{±65.17} & \textcolor{black}{0.0488} & \textcolor{black}{[74.70, 153.42]} \\
FedBN \citep{li2021fedbn} & 83.15\textsubscript{±15.58} & 77.08\textsubscript{±15.90} & 73.64\textsubscript{±22.26} & 77.95\textsubscript{±17.92} & \textcolor{black}{0.0116} & \textcolor{black}{[74.71, 81.63]} & 45.93\textsubscript{±45.05} & 65.68\textsubscript{±63.65} & 82.80\textsubscript{±62.37} & 64.81\textsubscript{±57.02} & \textcolor{black}{0.0269} & \textcolor{black}{[61.65, 113.04]} \\
FedProx \citep{li2020fedProx} & 75.54\textsubscript{±18.42} & 70.17\textsubscript{±21.48} & 70.10\textsubscript{±25.08} & 71.94\textsubscript{±21.66} & \textcolor{black}{0.0086} & \textcolor{black}{[70.10, 75.54]} & 101.98\textsubscript{±55.69} & 111.89\textsubscript{±66.33} & 113.44\textsubscript{±68.01} & 109.10\textsubscript{±63.34} & \textcolor{black}{0.0008} & \textcolor{black}{[111.29, 153.78]} \\
HarmoFL \citep{jiang2022harmofl} & 74.23\textsubscript{±18.86} & 69.23\textsubscript{±21.78} & 69.66\textsubscript{±25.21} & 71.04\textsubscript{±21.95} & \textcolor{black}{0.0081} & \textcolor{black}{[69.23, 72.70]} & 67.20\textsubscript{±32.22} & 73.58\textsubscript{±40.26} & 73.18\textsubscript{±43.77} & 71.32\textsubscript{±38.75} & \textcolor{black}{0.0001} & \textcolor{black}{[73.58, 122.39]} \\
HistoFL \citep{lu2022histofl} & 78.06\textsubscript{±17.39} & 72.90\textsubscript{±19.85} & 75.19\textsubscript{±21.96} & 75.38\textsubscript{±19.73} & \textcolor{black}{0.0402} & \textcolor{black}{[72.90, 78.06]} & 71.05\textsubscript{±41.71} & 88.74\textsubscript{±70.51} & 82.14\textsubscript{±66.59} & 80.64\textsubscript{±59.60} & \textcolor{black}{0.0172} & \textcolor{black}{[80.36, 143.66]} \\
FedFA \citep{zhou2023fedfa} & 77.16\textsubscript{±17.92} & 71.54\textsubscript{±20.61} & 71.89\textsubscript{±23.82} & 73.53\textsubscript{±20.78} & \textcolor{black}{0.0142} & \textcolor{black}{[71.54, 77.16]} & 80.55\textsubscript{±49.43} & 101.44\textsubscript{±70.07} & 110.13\textsubscript{±74.09} & 97.37\textsubscript{±64.53} & \textcolor{black}{0.0105} & \textcolor{black}{[89.86, 150.48]} \\
FedHEAL \citep{chen2024fair} & 78.81\textsubscript{±17.02} & 74.28\textsubscript{±19.57} & 74.84\textsubscript{±22.01} & 75.98\textsubscript{±19.53} & \textcolor{black}{0.0230} & \textcolor{black}{[74.28, 78.81]} & 60.96\textsubscript{±39.53} & 82.73\textsubscript{±74.09} & 85.62\textsubscript{±72.96} & 76.43\textsubscript{±62.19} & \textcolor{black}{0.0352} & \textcolor{black}{[70.27, 136.87]} \\
\textbf{PathFL (Ours)} & \textbf{85.65}\textsubscript{±10.86} & \textbf{83.36}\textsubscript{±9.21} & \textbf{80.16}\textsubscript{±17.62} & \textbf{83.05}\textsubscript{±12.57} & \textcolor{black}{-} & \textcolor{black}{[80.16, 85.65]} & \textbf{37.89}\textsubscript{±23.65} & \textbf{43.84}\textsubscript{±44.77} & \textbf{42.89}\textsubscript{±33.30} & \textbf{41.54}\textsubscript{±33.91} & \textcolor{black}{-} & \textcolor{black}{[43.84, 91.78]} \\
\bottomrule  
\end{tabular}%
}
\end{table*}

\begin{table*}[!t]
\centering
\caption{Cross-scanner results across different models and centers. The bold indicates the best value.}
\setlength{\tabcolsep}{3.5pt} 
\label{tabel: result_cross_scanner} 
\renewcommand{\arraystretch}{1.2} 
\resizebox{\textwidth}{!}{%
\begin{tabular}{rllLLll||lllLLl}
\toprule
        & \multicolumn{6}{c||}{\textbf{Dice (\%) $\uparrow$}}                    & \multicolumn{6}{c}{\textbf{ASSD \textit{(pix.)} $\downarrow$}}  \\
\cmidrule(lr){2-7} \cmidrule(lr){8-13}        
        & C1            & C2           & C3           & Average      & \textcolor{black}{p-value} & \textcolor{black}{95\% CI} & C1           & C2            & C3           & Average      & \textcolor{black}{p-value} & \textcolor{black}{95\% CI} \\
\midrule         
FedAvg \citep{McMahan2017fedavg} & 77.59\textsubscript{±18.65} & 80.31\textsubscript{±20.72} & 72.78\textsubscript{±24.25} & 76.90\textsubscript{±21.21} & \textcolor{black}{0.0127} & \textcolor{black}{[74.66, 79.46]} & 59.23\textsubscript{±38.42} & 80.53\textsubscript{±76.31} & 89.79\textsubscript{±71.98} & 76.51\textsubscript{±62.24} & \textcolor{black}{0.0088} & \textcolor{black}{[123.58, 170.89]} \\
FedBN \citep{li2021fedbn} & 77.60\textsubscript{±17.73} & 84.52\textsubscript{±14.18} & 77.35\textsubscript{±18.61} & 79.82\textsubscript{±16.84} & \textcolor{black}{0.0232} & \textcolor{black}{[77.47, 82.79]} & 52.05\textsubscript{±35.54} & 28.90\textsubscript{±23.91} & 60.80\textsubscript{±64.70} & 47.25\textsubscript{±41.38} & \textcolor{black}{0.0784} & \textcolor{black}{[109.92, 128.69]} \\
FedProx \citep{li2020fedProx} & 75.33\textsubscript{±20.21} & 79.26\textsubscript{±22.62} & 69.46\textsubscript{±26.26} & 74.68\textsubscript{±23.03} & \textcolor{black}{0.0114} & \textcolor{black}{[70.93, 78.12]} & 86.48\textsubscript{±51.52} & 97.69\textsubscript{±80.05} & 126.61\textsubscript{±69.25} & 103.59\textsubscript{±66.94} & \textcolor{black}{0.0066} & \textcolor{black}{[144.37, 194.54]} \\
HarmoFL \citep{jiang2022harmofl} & 74.79\textsubscript{±20.27} & 79.10\textsubscript{±22.87} & 69.25\textsubscript{±26.32} & 74.38\textsubscript{±23.15} & \textcolor{black}{0.0098} & \textcolor{black}{[70.64, 77.92]} & 54.73\textsubscript{±31.15} & 50.54\textsubscript{±46.21} & 73.82\textsubscript{±45.23} & 59.69\textsubscript{±40.86} & \textcolor{black}{0.0025} & \textcolor{black}{[124.49, 162.58]} \\
HistoFL \citep{lu2022histofl} & 76.11\textsubscript{±19.55} & 80.36\textsubscript{±21.38} & 72.35\textsubscript{±24.73} & 76.27\textsubscript{±21.88} & \textcolor{black}{0.0075} & \textcolor{black}{[73.33, 79.30]} & 76.15\textsubscript{±51.59} & 72.34\textsubscript{±69.53} & 75.01\textsubscript{±54.81} & 74.50\textsubscript{±58.64} & \textcolor{black}{0.0003} & \textcolor{black}{[125.48, 164.95]} \\
FedFA \citep{zhou2023fedfa} & 75.45\textsubscript{±19.84} & 79.89\textsubscript{±22.14} & 70.43\textsubscript{±25.76} & 75.25\textsubscript{±22.58} & \textcolor{black}{0.0106} & \textcolor{black}{[71.68, 78.73]} & 85.36\textsubscript{±52.58} & 75.31\textsubscript{±70.48} & 93.52\textsubscript{±65.66} & 84.73\textsubscript{±62.91} & \textcolor{black}{0.0009} & \textcolor{black}{[138.71, 167.64]} \\
FedHEAL \citep{chen2024fair} & 76.62\textsubscript{±19.19} & 80.62\textsubscript{±21.00} & 73.28\textsubscript{±24.15} & 76.84\textsubscript{±21.45} & \textcolor{black}{0.0098} & \textcolor{black}{[73.00, 79.45]} & 74.48\textsubscript{±51.68} & 68.47\textsubscript{±71.45} & 88.56\textsubscript{±74.99} & 77.17\textsubscript{±66.04} & \textcolor{black}{0.0019} & \textcolor{black}{[137.69, 170.04]} \\
\textbf{PathFL (Ours)}  & \textbf{81.59}\textsubscript{±14.06} & \textbf{85.77}\textsubscript{±11.34} & \textbf{82.38}\textsubscript{±14.90} & \textbf{83.25}\textsubscript{±13.43} & \textcolor{black}{-} & \textcolor{black}{[81.99, 85.14]} & \textbf{32.59}\textsubscript{±22.79} & \textbf{26.57}\textsubscript{±23.31} & \textbf{42.06}\textsubscript{±33.82}  & \textbf{33.74}\textsubscript{±26.64} & \textcolor{black}{-} & \textcolor{black}{[87.76, 124.40]} \\
\bottomrule  
\end{tabular}%
}
\end{table*}

\subsubsection{Performance on cross-modality datasets}
Our PathFL achieves superior performance on the cross-modality dataset with the highest average Dice of 88.93\% and the lowest average ASSD of 4.93, greatly outperforming the baseline (FedAvg) by 6.13\% in Dice and 2.50 in ASSD. The quantitative segmentation results on the cross-modality datasets for three centers (C1 to C3) are summarized in Table \ref{tabel: result_cross_modality}. Unlike the cross-source dataset, which consists of single modality H\&E-stained pathology images, the cross-modality dataset addresses challenges arising from diverse imaging modalities of cytology, fluorescence, and histology. Our model effectively handles cross-modality heterogeneity by leveraging similarities in cellular patterns, which remain consistent across different modalities. Other methods designed for single-modality heterogeneity often fail to adequately address these complexities, resulting in mutual degradation among client models. While our PathFL effectively enhances performance through its triple heterogeneity balancing strategy, which takes into account variations in images, features, and model aggregation. Specifically, PathFL shows substantial enhancements over FedAvg, with Dice increases of 6.52\% in C1, 2.42\% in C2, and 9.47\% in C3. These results highlight PathFL's capability to effectively address the challenges presented by diverse imaging modalities and maintain consistent performance.

\begin{table}[!t]

\caption{{ Unified Datasets Results across different models and clients. The bold indicates the best value.}}
\label{tab:table_unified}
\renewcommand{\arraystretch}{1.4}
\resizebox{\textwidth}{!}
{
\begin{tabular}{llllllllllllllllllll}
\hline
\textbf{} & \multicolumn{19}{c}{\textbf{Dice (\%) $\uparrow$}} \\ \cline{2-20} 
\textbf{} & C1 & C2 & C3 & C4 & C5 & C6 & C7 & C8 & C9 & C10 & C11 & C12 & C13 & C14 & C15 & C16 & Average & p-value & 95\% CI \\ \hline
FedAvg & 79.95 & 86.33 & 81.32 & 80.11 & 84.31 & 80.33 & 75.90 & 77.04 & 89.86 & 67.82 & 78.71 & 71.25 & 71.49 & 76.46 & 81.03 & 73.87 & 78.49 & 1.123E-05 & {[}75.85, 81.30{]} \\
  & \textsuperscript{±3.51} & \textsuperscript{±3.47} & \textsuperscript{±10.40} & \textsuperscript{±1.35} & \textsuperscript{±9.80} & \textsuperscript{±3.30} & \textsuperscript{±8.56} & \textsuperscript{±10.26} & \textsuperscript{±5.64} & \textsuperscript{±7.97} & \textsuperscript{±16.61} & \textsuperscript{±19.97} & \textsuperscript{±23.30} & \textsuperscript{±18.68} & \textsuperscript{±20.27} & \textsuperscript{±23.18} & \textsuperscript{±11.64} &  & \\
FedBN & 81.31 & 86.87 & 82.06 & 81.74 & \textbf{87.07} & 79.49 & 79.63 & 83.45 & 88.53 & 70.57 & 76.87 & 73.12 & 70.95 & 75.54 & 79.88 & 72.51 & 79.35 & 2.501E-04 & {[}76.55, 81.95{]} \\
 & \textsuperscript{±2.72} & \textsuperscript{±3.64} & \textsuperscript{±6.33} & \textsuperscript{±4.17} & \textsuperscript{±7.01} & \textsuperscript{±4.70} & \textsuperscript{±4.86} & \textsuperscript{±4.09} & \textsuperscript{±6.00} & \textsuperscript{±6.77} & \textsuperscript{±18.20} & \textsuperscript{±19.33} & \textsuperscript{±24.24} & \textsuperscript{±19.75} & \textsuperscript{±21.26} & \textsuperscript{±23.61} & \textsuperscript{±11.04} &  &  \\
FedProx & 80.11 & 86.33 & 81.22 & 80.59 & 84.69 & 80.67 & 77.00 & 70.84 & 88.55 & 66.13 & 81.07 & 70.57 & 72.21 & 75.72 & 80.42 & 73.61 & 78.11 & 5.084E-04 & {[}75.27, 80.98{]} \\
 & \textsuperscript{±3.39} & \textsuperscript{±3.35} & \textsuperscript{±9.30} & \textsuperscript{±2.85} & \textsuperscript{±8.86} & \textsuperscript{±3.36} & \textsuperscript{±4.81} & \textsuperscript{±13.27} & \textsuperscript{±5.71} & \textsuperscript{±8.41} & \textsuperscript{±15.92} & \textsuperscript{±20.63} & \textsuperscript{±22.67} & \textsuperscript{±19.42} & \textsuperscript{±20.88} & \textsuperscript{±23.45} & \textsuperscript{±11.64} &  &  \\
HarmoFL & 78.78 & 86.16 & 78.82 & 81.17 & 85.02 & 80.67 & 75.95 & 73.69 & 87.02 & 64.49 & 74.35 & 69.24 & 69.66 & 74.90 & 79.49 & 69.34 & 76.80 & 3.382E-05 & {[}73.44, 79.97{]} \\
 & \textsuperscript{±3.98} & \textsuperscript{±3.13} & \textsuperscript{±9.04} & \textsuperscript{±5.36} & \textsuperscript{±8.77} & \textsuperscript{±2.23} & \textsuperscript{±7.89} & \textsuperscript{±8.98} & \textsuperscript{±6.99} & \textsuperscript{±8.82} & \textsuperscript{±18.82} & \textsuperscript{±21.77} & \textsuperscript{±25.21} & \textsuperscript{±20.17} & \textsuperscript{±22.08} & \textsuperscript{±26.27} & \textsuperscript{±12.47} &  &  \\
HistoFL & 81.19 & 86.78 & 81.73 & 81.76 & 86.37 & 78.87 & \textbf{80.65} & 83.09 & 88.10 & 72.11 & 78.48 & 72.68 & 71.05 & 75.40 & 79.98 & 72.03 & 79.39 & 2.648E-04 & {[}76.74, 81.91{]} \\
 & \textsuperscript{±2.55} & \textsuperscript{±3.94} & \textsuperscript{±6.03} & \textsuperscript{±1.08} & \textsuperscript{±7.15} & \textsuperscript{±4.90} & \textsuperscript{±4.75} & \textsuperscript{±4.51} & \textsuperscript{±6.45} & \textsuperscript{±6.43} & \textsuperscript{±17.13} & \textsuperscript{±19.47} & \textsuperscript{±24.60} & \textsuperscript{±19.44} & \textsuperscript{±20.51} & \textsuperscript{±23.86} & \textsuperscript{±10.80} &  &  \\
FedFA & 79.86 & 86.51 & 81.27 & 80.97 & 84.94 & 80.24 & 74.47 & 73.04 & 89.05 & 66.95 & 77.92 & 69.90 & 72.16 & 75.89 & 80.63 & 72.49 & 77.89 & 8.762E-05 & {[}75.25, 80.81{]} \\
 & \textsuperscript{±3.16} & \textsuperscript{±3.36} & \textsuperscript{±9.22} & \textsuperscript{±3.83} & \textsuperscript{±8.74} & \textsuperscript{±3.73} & \textsuperscript{±12.32} & \textsuperscript{±13.23} & \textsuperscript{±6.80} & \textsuperscript{±7.65} & \textsuperscript{±16.29} & \textsuperscript{±20.80} & \textsuperscript{±22.44} & \textsuperscript{±19.04} & \textsuperscript{±20.87} & \textsuperscript{±24.14} & \textsuperscript{±12.23} &  &  \\
FedHEAL & 76.33 & 84.96 & 56.06 & 79.67 & 80.25 & 70.49 & 63.55 & 84.02 & 89.51 & 73.27 & 74.23 & 69.22 & 69.66 & 74.80 & 79.17 & 69.25 & 74.65 & 4.127E-04 & {[}70.82, 78.63{]} \\
 & \textsuperscript{±4.36} & \textsuperscript{±3.70} & \textsuperscript{±20.61} & \textsuperscript{±3.50} & \textsuperscript{±9.49} & \textsuperscript{±4.63} & \textsuperscript{±11.89} & \textsuperscript{±5.88} & \textsuperscript{±5.99} & \textsuperscript{±6.66} & \textsuperscript{±18.87} & \textsuperscript{±21.79} & \textsuperscript{±25.21} & \textsuperscript{±20.27} & \textsuperscript{±22.85} & \textsuperscript{±26.32} & \textsuperscript{±13.25} &  &  \\
\textbf{PathFL} & \textbf{81.86} & \textbf{87.08} & \textbf{85.72} & \textbf{84.02} & 86.71 & \textbf{81.09} & 78.48 & \textbf{87.25} & \textbf{93.00} & \textbf{74.67} & \textbf{84.00} & \textbf{75.33} & \textbf{73.79} & \textbf{79.85} & \textbf{85.33} & \textbf{80.50} & \textbf{82.42} & - & {[}79.95, 84.97{]} \\
\textbf{(Ours)} & \textsuperscript{±3.41} & \textsuperscript{±3.69} & \textsuperscript{±7.48} & \textsuperscript{±3.91} & \textsuperscript{±9.38} & \textsuperscript{±2.27} & \textsuperscript{±5.69} & \textsuperscript{±3.88} & \textsuperscript{±4.28} & \textsuperscript{±6.65} & \textsuperscript{±12.44} & \textsuperscript{±18.24} & \textsuperscript{±22.10} & \textsuperscript{±16.40} & \textsuperscript{±15.43} & \textsuperscript{±20.35} & \textsuperscript{±9.72} &  &  \\ \hline \hline

 & \multicolumn{19}{c}{\textbf{ASSD \textit{(pix.)} $\downarrow$}} \\
 \cline{2-20} 
 & C1 & C2 & C3 & C4 & C5 & C6 & C7 & C8 & C9 & C10 & C11 & C12 & C13 & C14 & C15 & C16 & Average & p-value & 95\% CI \\ \hline
FedAvg & 2.88 & 2.46 & 51.45 & 2.91 & 19.72 & 3.29 & 7.20 & 10.53 & 14.32 & 12.48 & 88.15 & 109.28 & 153.95 & 128.16 & 178.28 & 132.72 & 57.36 & 2.921E-03 & {[}30.21, 89.83{]} \\
 & \textsuperscript{±0.67} & \textsuperscript{±0.93} & \textsuperscript{±40.47} & \textsuperscript{±0.22} & \textsuperscript{±18.56} & \textsuperscript{±0.77} & \textsuperscript{±6.38} & \textsuperscript{±8.51} & \textsuperscript{±16.81} & \textsuperscript{±5.66} & \textsuperscript{±88.44} & \textsuperscript{±68.32} & \textsuperscript{±95.85} & \textsuperscript{±98.02} & \textsuperscript{±131.17} & \textsuperscript{±119.74} & \textsuperscript{±43.78} &  &  \\
FedBN & 2.97 & \textbf{2.40} & \textbf{39.75} & 2.39 & 16.35 & 3.53 & \textbf{4.74} & 3.52 & 15.34 & 11.13 & 95.47 & 104.57 & 130.96 & 146.06 & 192.55 & 143.66 & 57.21 & 3.136E-02 & {[}27.33, 87.52{]} \\
 & \textsuperscript{±0.70} & \textsuperscript{±0.84} & \textsuperscript{±22.45} & \textsuperscript{±0.50} & \textsuperscript{±9.86} & \textsuperscript{±0.86} & \textsuperscript{±3.94} & \textsuperscript{±2.43} & \textsuperscript{±13.82} & \textsuperscript{±4.37} & \textsuperscript{±92.16} & \textsuperscript{±69.03} & \textsuperscript{±119.72} & \textsuperscript{±121.46} & \textsuperscript{±133.90} & \textsuperscript{±116.79} & \textsuperscript{±44.55} &  &  \\
FedProx & 2.94 & 2.45 & 49.45 & 2.72 & 21.03 & 3.21 & 7.56 & 11.92 & 18.89 & 13.85 & 75.03 & 109.38 & 192.30 & 139.77 & 175.17 & 133.71 & 59.96 & 7.463E-03 & {[}29.36, 92.36{]} \\
 & \textsuperscript{±0.68} & \textsuperscript{±0.92} & \textsuperscript{±45.68} & \textsuperscript{±0.20} & \textsuperscript{±19.28} & \textsuperscript{±0.74} & \textsuperscript{±6.97} & \textsuperscript{±8.02} & \textsuperscript{±19.15} & \textsuperscript{±6.03} & \textsuperscript{±91.60} & \textsuperscript{±59.95} & \textsuperscript{±174.70} & \textsuperscript{±111.52} & \textsuperscript{±126.68} & \textsuperscript{±121.20} & \textsuperscript{±49.58} &  &  \\
HarmoFL & 3.86 & 2.49 & 40.80 & 2.59 & 18.43 & \textbf{3.15} & 9.69 & 10.02 & 19.48 & 14.33 & 74.42 & \textbf{75.25} & \textbf{119.92} & 118.67 & 179.01 & 131.76 & 51.49 & 4.856E-01 & {[}26.58, 79.96{]} \\
 & \textsuperscript{±2.61} & \textsuperscript{±0.94} & \textsuperscript{±32.44} & \textsuperscript{±0.64} & \textsuperscript{±16.51} & \textsuperscript{±0.48} & \textsuperscript{±11.78} & \textsuperscript{±7.83} & \textsuperscript{±15.39} & \textsuperscript{±6.62} & \textsuperscript{±66.09} & \textsuperscript{±39.94} & \textsuperscript{±114.21} & \textsuperscript{±131.32} & \textsuperscript{±171.61} & \textsuperscript{±121.47} & \textsuperscript{±46.24} &  &  \\
HistoFL & 2.91 & 2.45 & 46.42 & 2.89 & 17.38 & 3.69 & 4.78 & 4.21 & 17.20 & 10.18 & 85.04 & 86.83 & 147.07 & 151.82 & 189.73 & 143.72 & 57.27 & 6.309E-03 & {[}28.07, 88.12{]} \\
 & \textsuperscript{±0.67} & \textsuperscript{±0.86} & \textsuperscript{±35.51} & \textsuperscript{±0.54} & \textsuperscript{±9.31} & \textsuperscript{±0.80} & \textsuperscript{±3.72} & \textsuperscript{±3.29} & \textsuperscript{±15.28} & \textsuperscript{±3.85} & \textsuperscript{±90.75} & \textsuperscript{±68.73} & \textsuperscript{±148.42} & \textsuperscript{±118.01} & \textsuperscript{±132.76} & \textsuperscript{±116.87} & \textsuperscript{±46.84} &  &  \\
FedFA & 3.15 & 2.44 & 48.53 & 2.64 & 20.68 & 3.30 & 13.21 & 11.08 & 13.77 & 13.19 & 97.10 & 120.36 & 189.23 & 133.46 & 173.08 & 140.20 & 61.59 & 6.177E-03 & {[}33.02, 94.01{]} \\
 & \textsuperscript{±0.77} & \textsuperscript{±0.89} & \textsuperscript{±39.26} & \textsuperscript{±0.39} & \textsuperscript{±19.31} & \textsuperscript{±0.80} & \textsuperscript{±21.93} & \textsuperscript{±8.67} & \textsuperscript{±15.74} & \textsuperscript{±6.03} & \textsuperscript{±70.57} & \textsuperscript{±61.05} & \textsuperscript{±189.27} & \textsuperscript{±98.70} & \textsuperscript{±132.82} & \textsuperscript{±116.22} & \textsuperscript{±48.90} &  &  \\
FedHEAL & 4.05 & 2.67 & 126.38 & 2.83 & 21.67 & 5.66 & 8.03 & 6.19 & 16.23 & 9.40 & 151.34 & 145.48 & 175.42 & 166.40 & 196.62 & 180.55 & 76.18 & 4.615E-03 & {[}38.44, 116.08{]} \\
 & \textsuperscript{±1.68} & \textsuperscript{±0.95} & \textsuperscript{±47.86} & \textsuperscript{±0.21} & \textsuperscript{±17.41} & \textsuperscript{±1.45} & \textsuperscript{±4.49} & \textsuperscript{±7.38} & \textsuperscript{±15.81} & \textsuperscript{±4.57} & \textsuperscript{±54.43} & \textsuperscript{±55.26} & \textsuperscript{±81.10} & \textsuperscript{±90.73} & \textsuperscript{±110.23} & \textsuperscript{±80.90} & \textsuperscript{±35.91} &  &  \\
\textbf{PathFL} & \textbf{2.80} & 2.44 & 39.91 & \textbf{2.16} & \textbf{16.01} & 3.31 & 6.84 & \textbf{2.81} & \textbf{6.68} & \textbf{9.31} & \textbf{68.51} & 79.32 & 143.34 & \textbf{126.61} & \textbf{169.98} & \textbf{119.12} & \textbf{49.95} & - & {[}22.11, 77.47{]} \\
\textbf{(Ours)} & \textsuperscript{±0.71} & \textsuperscript{±1.17} & \textsuperscript{±40.21} & \textsuperscript{±0.40} & \textsuperscript{±13.53} & \textsuperscript{±0.80} & \textsuperscript{±6.23} & \textsuperscript{±2.08} & \textsuperscript{±9.49} & \textsuperscript{±4.53} & \textsuperscript{±107.13} & \textsuperscript{±70.44} & \textsuperscript{±131.87} & \textsuperscript{±92.20} & \textsuperscript{±121.82} & \textsuperscript{±78.13} & \textsuperscript{±30.05} &  &  \\ \hline
\end{tabular}
}
\end{table}

\subsubsection{Performance on cross-organ datasets}
Our PathFL achieves superior performance on the cross-organ dataset with the highest average Dice of 83.05\% and the lowest average ASSD of 41.54, greatly outperforming FedAvg by 7.94\% in Dice and 40.12 in ASSD. The quantitative segmentation results on the cross-organ datasets for three centers (C1 to C3) are summarized in Table \ref{tabel: result_cross_organ}. The cross-organ dataset focuses on a consistent adenocarcinoma segmentation task across various organs, featuring three distinct adenocarcinomas, all acquired from the same scanner. This uniformity allows for a clearer comparison of model performance across different anatomical contexts. The results demonstrate that our method consistently outperforms other SOTAs, achieving Dice of 85.65\% at C1, 83.36\% at C2, and 80.16\% at C3. Moreover, PathFL significantly reduces ASSD compared to the baseline, with decreases of 27.50 at C1, 43.77 at C2, and 49.09 at C3, highlighting its enhanced boundary accuracy and segmentation consistency. In contrast, other approaches like FedProx and FedFA, exhibit limitations in effectively addressing this cross-organ variability, with average ASSD of 109.10 and 97.37, respectively. These methods often struggle to fully account for substantial density variations and anatomical differences across diverse datasets, consequently compromising segmentation performance. PathFL uniquely mitigates these heterogeneity challenges through a comprehensive, multi-perspective approach, substantially enhancing segmentation robustness and accuracy.

\subsubsection{Performance on cross-scanner datasets}
Our PathFL achieves superior performance on the cross-scanner dataset with the highest average Dice of 83.25\% and the lowest average ASSD of 33.74, greatly outperforming FedAvg by 6.35\% in Dice and 42.77 in ASSD. The quantitative segmentation results on the cross-scanner datasets for three centers (C1 to C3) are summarized in Table \ref{tabel: result_cross_scanner}. The cross-scanner dataset illustrates the complexities inherent in segmentation tasks when utilizing images acquired from various scanners, which may differ in resolution, contrast, and noise characteristics. Our method not only surpasses other SOTAs but also demonstrates consistent performance across clients, achieving Dice of 81.59\% at C1, 85.77\% at C2, and 82.28\% at C3. The findings from the cross-scanner dataset affirm the generalization of our method, illustrating its capacity to maintain high segmentation accuracy over scanner diversity. This reinforces the potential of our method for real-world applications, where the integration of multi-source imaging data is increasingly prevalent, and emphasizes its applicability in enhancing segmentation precision in clinical settings.

{ 
\subsubsection{Performance on unified datasets}
Our PathFL achieves the best segmentation performance, yielding a substantial and consistent improvement across all 16 clients as Table \ref{tab:table_unified} shown. Specifically, our PathFL significantly outperforms other methods on the unified dataset, attaining an average Dice of 82.42\% and an average ASSD of 49.95. PathFL also exhibits significant superiority over the baseline (FedAvg), with an average improvement of 3.93\% in Dice and 7.41 in ASSD. Other methods like HistoFL perform well on few clients but lack generalization, leading to reduced overall performance. In contrast, PathFL achieves a superior balance between accuracy and stability, consistently maintaining high Dice with lower variance. Furthermore, statistical tests confirm the statistical significance of our PathFL’s segmentation improvements, with most p-values below 0.01. The 95\% CIs for Dice ([79.95, 84.97]) and ASSD ([22.11, 77.47]) demonstrate superior consistency and robustness across clients compared to other methods. In summary, these findings collectively confirm that PathFL not only advances federated segmentation in highly heterogeneous scenarios but also ensures stable and reliable performance across diverse datasets, making it a promising solution for real-world medical applications.
}

\begin{figure*}[htp]
\centering
\includegraphics[width=\textwidth]{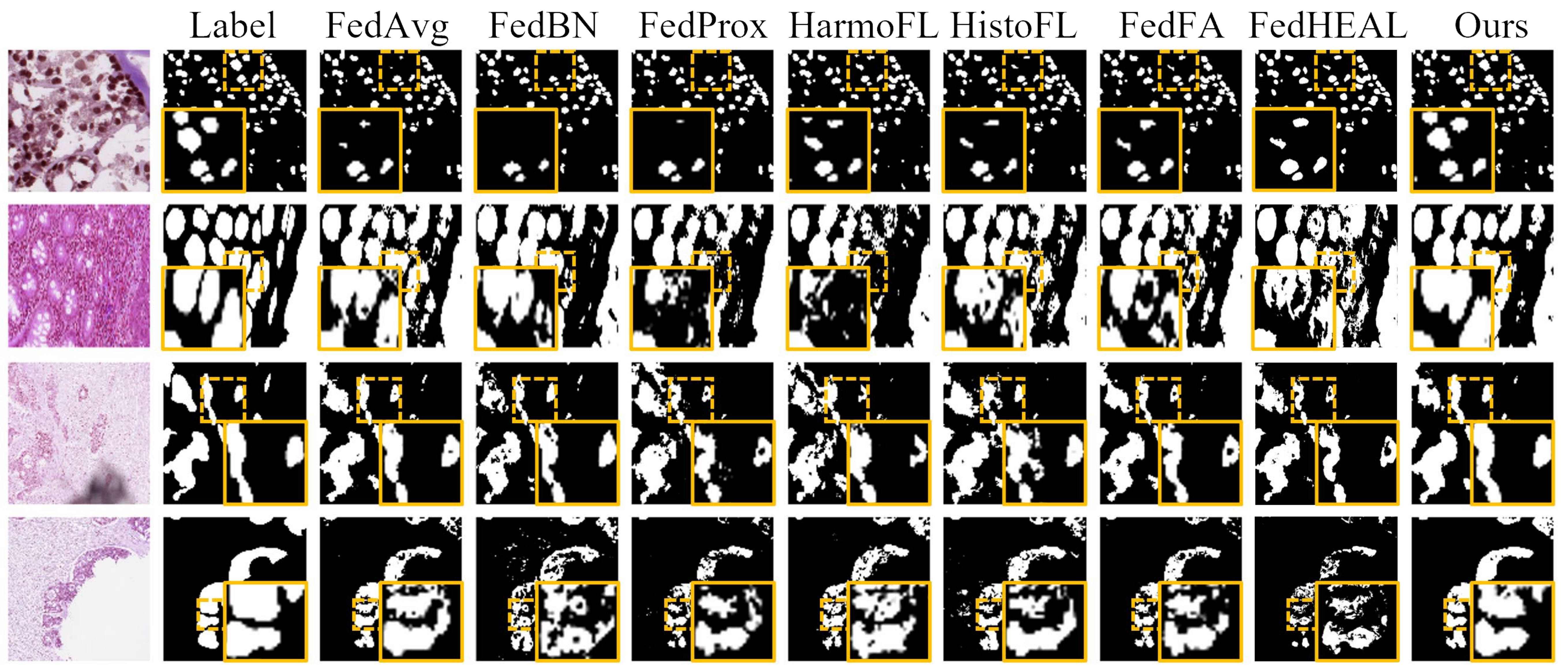}
\caption{Qualitative visualization comparison on segmentation results with our method and other state-of-the-art methods. From left to right: (a) Original Image, (b) Label, (c) FedAvg, (d) FedBN, (e) FedProx, (f) HarmoFL, (g) HistoFL, (h) FedFA, (i) FedHEAL, and (j) PathFL. Our PathFL achieves improved boundary accuracy and structure preservation compared to other methods.}
\label{fig_VISUAL}
\end{figure*}

\subsubsection{Visualization results}
The visualization results (see Fig.~\ref{fig_VISUAL}) demonstrate the segmentation performance of our model, showcasing its ability to accurately delineate regions of interest across diverse pathology images. Our segmentation results reveal the ability to identify cell nuclei that other models fail to recognize, effectively reducing segmentation voids and minimizing fragmented regions. This improvement enhances the completeness of the segmentation and significantly increases its accuracy. The results highlight the effectiveness of our PathFL in handling multi-center heterogeneity, with clear segmentation boundaries and improved consistency across different datasets. These visualizations emphasize the model's robustness in capturing both structural and stylistic features, reinforcing its exceptional segmentation.

\begin{table}[]
\caption{Ablation analysis of modules on cross-source datasets. $\Delta$ denotes the Dice improvement compared to the baseline.}
\label{tab:table_ablation_small}
\resizebox{0.4\linewidth}{!}
{
\begin{tabular}{lll||cc}
\toprule
CSE & AFA & SSA & Average Dice (\%) $\uparrow$ & {$\Delta$} \\ 
\midrule
 &  &  & 82.62\textsubscript{±4.99} & - \\
\checkmark &  &  & 83.67\textsubscript{±4.25} & +1.05 \\
 & \checkmark &  & 83.43\textsubscript{±4.61} & +0.81 \\
 &  & \checkmark & 83.33\textsubscript{±4.67} & +0.71 \\
\checkmark & \checkmark &  & 83.94\textsubscript{±4.84} & +1.32 \\
\checkmark &  & \checkmark & 83.81\textsubscript{±4.91} & +1.19 \\
 & \checkmark & \checkmark & 83.48\textsubscript{±4.39} & +0.86 \\
\textbf{\checkmark} & \textbf{\checkmark} & \textbf{\checkmark} & \textbf{84.27\textsubscript{±4.43}} & \textbf{+1.65} \\ 
\bottomrule
\end{tabular}
}
\end{table}

\subsubsection{Ablation study}
We conducted a comprehensive ablation study on cross-source datasets to systematically evaluate the performance contributions of our proposed modules, as detailed in Table \ref{tab:table_ablation_small}. We explored various combinations of the three modules: the image-level module (Collaborative Style Enhancement, CSE), the feature-level module (Adaptive Feature Alignment, AFA), and the model aggregation-level module (Stratified Similarity Aggregation, SSA), with FedAvg serving as the baseline. Notably, our experimental results reveal that in the most challenging cross-source pathology segmentation scenarios, the image-level heterogeneity handling module demonstrates the most significant performance improvements. The ablation analysis further substantiates the synergistic interactions between these multi-level modules, underscoring the holistic design and effectiveness of our proposed framework.

\begin{table}[]

\caption{{Comparison of four backbones on PathFL and FedAvg for cross-source datasets. The bold indicates the best value. $\Delta$ indicates the improvement of PathFL over FedAvg with the same backbone.}}
\label{tab:table_foundation_crosssource}
\renewcommand{\arraystretch}{1.2}
\resizebox{\linewidth}{!}{
\begin{tabular}{llllllllll}
\hline
 & \multicolumn{8}{c}{\textbf{Dice (\%) $\uparrow$}} &  \\ \cline{2-10} 
 & C1 & C2 & C3 & C4 & C5 & C6 & C7 & Average & $\Delta$ \\ \hline
FedAvg+U-Net & 80.48\textsubscript{±3.82} & 87.01\textsubscript{±3.25} & 84.09\textsubscript{±8.15} & 81.81\textsubscript{±2.26} & 86.97\textsubscript{±8.21} & 79.99\textsubscript{±3.29} & 78.01\textsubscript{±5.96} & 82.62\textsubscript{±4.99} &  \\
PathFL+U-Net & 82.06\textsubscript{±3.57} & \textbf{87.46}\textsubscript{±3.24} & 86.12\textsubscript{±5.02} & 84.72\textsubscript{±3.44} & 88.96\textsubscript{±8.27} & 81.44\textsubscript{±2.75} & \textbf{79.17}\textsubscript{±4.70} & 84.27\textsubscript{±4.43} & 1.65$\uparrow$ \\
FedAvg+ViT & 69.24\textsubscript{±6.30} & 83.15\textsubscript{±3.97} & 72.94\textsubscript{±12.20} & 75.16\textsubscript{±4.71} & 75.07\textsubscript{±8.43} & 74.14\textsubscript{±4.91} & 63.44\textsubscript{±10.59} & 73.31\textsubscript{±7.30} &  \\
PathFL+ViT & 74.10\textsubscript{±4.88} & 83.37\textsubscript{±4.08} & 68.35\textsubscript{±14.61} & 78.13\textsubscript{±3.99} & 74.33\textsubscript{±10.57} & 75.88\textsubscript{±3.39} & 68.16\textsubscript{±9.02} & 74.62\textsubscript{±7.22} & 1.31$\uparrow$ \\
FedAvg+CHIEF & 78.22\textsubscript{±4.07} & 85.35\textsubscript{±4.02} & 76.38\textsubscript{±10.45} & 80.20\textsubscript{±4.27} & 83.19\textsubscript{±10.59} & 79.29\textsubscript{±3.55} & 75.48\textsubscript{±7.68} & 79.73\textsubscript{±6.38} &  \\
PathFL+CHIEF & 80.15\textsubscript{±3.13} & 86.64\textsubscript{±4.06} & 77.27\textsubscript{±9.95} & 82.05\textsubscript{±5.40} & 88.63\textsubscript{±7.02} & 77.55\textsubscript{±5.71} & 77.73\textsubscript{±5.78} & 81.43\textsubscript{±5.86} & 1.7$\uparrow$ \\
FedAvg+CONCH & 80.13\textsubscript{±3.57} & 80.08\textsubscript{±4.07} & 90.39\textsubscript{±4.77} & 82.79\textsubscript{±3.83} & 91.74\textsubscript{±6.07} & 77.22\textsubscript{±3.25} & 72.37\textsubscript{±4.87} & 82.10\textsubscript{±4.35} &  \\
PathFL+CONCH & \textbf{82.45}\textsubscript{±3.03} & 83.33\textsubscript{±3.64} & \textbf{91.00}\textsubscript{±4.77} & \textbf{85.09}\textsubscript{±3.55} & \textbf{92.16}\textsubscript{±6.07} & \textbf{81.63}\textsubscript{±2.42} & 75.01\textsubscript{±3.74} & \textbf{84.38}\textsubscript{±3.89} & 2.28$\uparrow$ \\ \hline
\end{tabular}
}
\end{table}

\begin{table}[]

\caption{{Comparison of four backbones on PathFL and FedAvg for three datasets. The bold indicates the best value. $\Delta$ indicates the improvement of PathFL over FedAvg with the same backbone.}}
\label{tab:table_foundation_3datasets}
\renewcommand{\arraystretch}{1.1}
\resizebox{0.6\linewidth}{!}{
\begin{tabular}{llllll}
\hline
 & \multicolumn{5}{c}{\textbf{Cross-modality Datasets Dice (\%) $\uparrow$}} \\ \cline{2-6} 
 & C1 & C2 & C3 & Average & $\Delta$ \\ \hline
FedAvg+U-Net & 83.41\textsubscript{±9.85} & 91.46\textsubscript{±4.64} & 73.53\textsubscript{±6.71} & 82.80\textsubscript{±7.07} &  \\
PathFL+U-Net & 89.93\textsubscript{±3.49} & 93.88\textsubscript{±3.10} & 83.00\textsubscript{±5.30} & 88.93\textsubscript{±3.96} & 6.13$\uparrow$ \\
FedAvg+ViT & 68.21\textsubscript{±10.26} & 89.59\textsubscript{±6.01} & 60.13\textsubscript{±8.23} & 72.65\textsubscript{±8.17} &  \\
PathFL+ViT & 89.32\textsubscript{±3.08} & 95.00\textsubscript{±3.74} & 81.56\textsubscript{±5.24} & 88.63\textsubscript{±4.02} & 15.98$\uparrow$ \\
FedAvg+CHIEF & 82.58\textsubscript{±5.01} & 88.15\textsubscript{±6.03} & 66.86\textsubscript{±6.88} & 79.20\textsubscript{±5.97} &  \\
PathFL+CHIEF & \textbf{92.90}\textsubscript{±2.96} & \textbf{94.45}\textsubscript{±4.45} & 83.71\textsubscript{±5.48} & 90.35\textsubscript{±4.30} & 11.15$\uparrow$ \\
FedAvg+CONCH & 84.04\textsubscript{±4.15} & 91.61\textsubscript{±4.40} & 72.93\textsubscript{±5.63} & 82.86\textsubscript{±4.72} &  \\
PathFL+CONCH & 91.60\textsubscript{±3.88} & 94.20\textsubscript{±3.38} & \textbf{85.75}\textsubscript{±3.54} & \textbf{90.52}\textsubscript{±3.60} & 7.56$\uparrow$ \\ \hline
 & \multicolumn{5}{c}{\textbf{Cross-oragn Datasets Dice (\%) $\uparrow$}} \\ \cline{2-6} 
 & C1 & C2 & C3 & Average & $\Delta$ \\ \hline
FedAvg+U-Net & 79.08\textsubscript{±16.72} & 72.79\textsubscript{±20.83} & 73.47\textsubscript{±22.67} & 75.11\textsubscript{±20.07} &  \\
PathFL+U-Net & \textbf{85.65}\textsubscript{±10.86} & 83.36\textsubscript{±9.21} & 80.16\textsubscript{±17.62} & 83.05\textsubscript{±12.57} & 7.94$\uparrow$ \\
FedAvg+ViT & 74.87\textsubscript{±18.16} & 70.98\textsubscript{±20.80} & 71.56\textsubscript{±23.66} & 72.47\textsubscript{±20.88} &  \\
PathFL+ViT & 75.18\textsubscript{±9.31} & 74.15\textsubscript{±9.58} & 77.54\textsubscript{±12.97} & 75.62\textsubscript{±10.62} & 3.15$\uparrow$ \\
FedAvg+CHIEF & 76.89\textsubscript{±18.41} & 74.65\textsubscript{±19.65} & 72.45\textsubscript{±23.29} & 74.66\textsubscript{±20.45} &  \\
PathFL+CHIEF & 78.51\textsubscript{±18.18} & 69.22\textsubscript{±21.79} & 79.51\textsubscript{±18.11} & 75.74\textsubscript{±19.36} & 1.08$\uparrow$ \\
FedAvg+CONCH & 85.18\textsubscript{±13.34} & 84.18\textsubscript{±9.74} & 84.73\textsubscript{±13.95} & 84.70\textsubscript{±12.34} &  \\
PathFL+CONCH & 85.28\textsubscript{±12.36} & \textbf{86.96}\textsubscript{±8.27} & \textbf{86.06}\textsubscript{±13.08} & \textbf{86.10}\textsubscript{±11.24} & 1.40$\uparrow$ \\ \hline
 & \multicolumn{5}{c}{\textbf{Cross-scanner Datasets Dice (\%) $\uparrow$}} \\ \cline{2-6} 
 & C1 & C2 & C3 & Average & $\Delta$ \\ \hline
FedAvg+U-Net & 77.59\textsubscript{±18.65} & 80.31\textsubscript{±20.72} & 72.78\textsubscript{±24.25} & 76.90\textsubscript{±21.21} &  \\
PathFL+U-Net & 81.59\textsubscript{±14.06} & 85.77\textsubscript{±11.34} & 82.38\textsubscript{±14.90} & 83.25\textsubscript{±13.43} & 6.35$\uparrow$ \\
FedAvg+ViT & 75.10\textsubscript{±19.69} & 79.26\textsubscript{±22.25} & 70.23\textsubscript{±25.91} & 74.86\textsubscript{±22.62} &  \\
PathFL+ViT & 75.52\textsubscript{±19.00} & 79.10\textsubscript{±22.87} & 70.81\textsubscript{±25.14} & 75.14\textsubscript{±22.34} & 0.28$\uparrow$ \\
FedAvg+CHIEF & 75.83\textsubscript{±19.54} & 79.42\textsubscript{±21.93} & 72.18\textsubscript{±24.84} & 75.81\textsubscript{±22.10} &  \\
PathFL+CHIEF & 78.04\textsubscript{±17.49} & 79.03\textsubscript{±22.81} & 74.93\textsubscript{±22.03} & 77.33\textsubscript{±20.78} & 1.52$\uparrow$ \\
FedAvg+CONCH & 80.92\textsubscript{±13.91} & 85.51\textsubscript{±14.77} & 82.44\textsubscript{±16.81} & 82.96\textsubscript{±15.17} &  \\
PathFL+CONCH & \textbf{83.79}\textsubscript{±11.28} & \textbf{86.21}\textsubscript{±14.07} & \textbf{84.80}\textsubscript{±15.13} & \textbf{84.93}\textsubscript{±13.49} & 1.97$\uparrow$ \\ \hline
\end{tabular}
}
\end{table}

{ \subsubsection{Comparison with different backbones}
To further explore the impact of different SOTA backbones within our PathFL framework, we conducted additional experiments using Vision Transformer (ViT) \citep{dosovitskiy2020vit} and two pathology foundation models  (CONCH \citep{lu2024conch} and CHIEF \citep{wang2024chief}) alongside U-Net. Specifically, ViT was configured with six attention heads, eight depths, and an MLP dimension of 1536. For foundation models, we leveraged their pretrained encoders as frozen feature extractors coupled with an upsampling convolutional decoder. The results, summarized in Table \ref{tab:table_foundation_crosssource} and Table \ref{tab:table_foundation_3datasets}, show that our PathFL framework consistently outperforms the baseline (FedAvg) across all four backbones, confirming its broad applicability across different network architectures. PathFL with the CONCH backbone achieves the best performance across cross-source, cross-modality, cross-scanner, and cross-organ datasets, with average Dice scores of 84.38\%, 90.52\%, 86.10\%, and 84.93\%, respectively. The suboptimal performance of standalone foundation models with standard FedAvg in addressing cross-client heterogeneity stems from their inherent design prioritizing generalized feature extraction over local adaptability. Moreover, vision-language models such as CONCH and CHIEF predominantly focus on global semantic representations rather than local spatial details and limit their segmentation performance when confined to visual pathways alone. Our PathFL framework effectively bridges the gap between the global generalization capabilities of foundation models and the local adaptability required in FL. The integration of PathFL with foundation models like CONCH enables substantial performance gains, underscoring the effectiveness of our multi-alignment strategy in enhancing federated segmentation across heterogeneous data distributions.
}

%

\begin{table}[]
\caption{Computational cost per client of PathFL and FedAvg with different backbones on cross-source datasets. {(distributed) indicates distributed similarity computation, while the default is centralized; $\Delta$ denotes the difference between the distributed and the centralized computation.}}
\label{tab:table_computation}
\renewcommand{\arraystretch}{1.1}
\resizebox{\linewidth}{!}{
\begin{tabular}{llllllll}
\toprule
& \begin{tabular}[c]{@{}l@{}}FLOPs \\ (G)\end{tabular} & \begin{tabular}[c]{@{}l@{}}Params\\ (M)\end{tabular} & \begin{tabular}[c]{@{}l@{}}Average time \\ (ms)\end{tabular} & \begin{tabular}[c]{@{}l@{}}Communication overhead\\(MB)\end{tabular} & \begin{tabular}[c]{@{}l@{}} {$\Delta$ FLOPs} \\ {(G)}\end{tabular} & \begin{tabular}[c]{@{}l@{}} {$\Delta$ Average time} \\ {(ms)}\end{tabular} & \begin{tabular}[c]{@{}l@{}} {$\Delta$ Communication overhead}\\ {(MB)}\end{tabular} \\ \midrule

FedAvg+U-Net & 8.21 & 1.95 & 3.34 & 14.88 &  &  &  \\
PathFL+U-Net & 10.89 & 1.95 & 4.33 & 26.88 &  &  &  \\
{PathFL+UNet (distributed)} & {8.59} & {1.95} & {3.77} & {30.30} & {-2.30} & {-0.56} & {+3.42} \\
FedAvg+ViT & 7.98 & 15.12 & 4.31 & 115.39 &  &  &  \\
PathFL+ViT & 13.79 & 15.12 & 8.72 & 127.39 &  &  &  \\
{PathFL+VIT (distributed)}& {8.81} & {15.12} & {7.75} & {132.08} & {-4.98} & {-0.97} & {+4.69} \\
FedAvg+CHIEF & 15.87 & 91.84 & 14.59 & 702.86 &  &  &  \\
PathFL+CHIEF & 25.39 & 91.84 & 20.10 & 714.86 &  &  &  \\
{PathFL+CHIEF (distributed)} & {17.23} & {91.84} & {19.35} & {723.46} & {-8.16} & {-0.75} & {+8.60} \\
FedAvg+CONCH & 26.30 & 67.10 & 12.55 & 3527.56 &  &  &  \\
PathFL+CONCH & 38.96 & 67.10 & 21.47 & 3539.56 &  &  &  \\
{PathFL+CONCH (distributed)} & {28.11} & {67.10} & {18.36} & {3553.34} & {-10.85} & {-3.11} & {+13.78} \\ 
\bottomrule
\end{tabular}
}
\end{table}

 \subsubsection{Computational cost} To evaluate the computational cost of our PathFL framework when applied to larger backbones, such as ViT and foundation models like CONCH and CHIEF, we present key metrics in Table \ref{tab:table_computation}. These include FLOPs (billions, G), model parameters (millions, M), average training time per sample (milliseconds, ms), and communication overhead per client (MB). Compared with the baseline, the computation cost of our PathFL framework is comparable in all metrics. While the inclusion of the multi-alignment strategy slightly increases communication cost due to the exchange of style information between clients, the substantial improvements in segmentation performance justify this trade-off. As the backbone becomes more complex, the computational cost of PathFL increases. Thus, U-Net remains the most computationally efficient backbone, balancing segmentation accuracy and efficiency. {To further alleviate computational cost, especially when scaling to larger backbones, we evaluate a distributed similarity computation strategy. It shifts layer-wise similarity calculations from the server to clients, enabling parallel local computation based on the transmission of lightweight features between clients. This reduces local FLOPs and training time with a slight increase in communication overhead, thereby facilitating the scalability of PathFL to larger backbones.}

\section{Discussion and Conclusion}
{ Heterogeneity remains a fundamental challenge in federated learning for pathology segmentation, significantly impacting model performance and the feasibility of real-world deployment. To simulate multi-center collaborations, we explored four representative heterogeneous datasets capturing real-world pathology variations: data sources (cross-source), imaging techniques (cross-modality), anatomical regions (cross-organ), and scanning devices (cross-scanner). For instance, scenarios involving the same disease affecting different organs exemplify cross-organ heterogeneity, while variations in scanning technology for the same disease within the same organ illustrate cross-scanner heterogeneity. Our multi-alignment strategy demonstrates remarkable adaptability across heterogeneous scenarios. Experimental results demonstrate that PathFL with the U-Net backbone achieves an optimal balance between performance and efficiency, making it a lightweight yet effective choice for practical applications. Meanwhile, PathFL with the CONCH foundation model provides superior performance for all heterogeneous datasets, highlighting the potential of integrating foundation models into FL. Our findings suggest a promising direction for developing federated foundation models, capable of unlocking the potential of large-scale collaborative learning while preserving data privacy.
}

In conclusion, this study introduces PathFL, a novel multi-alignment federated learning framework that comprehensively addresses the critical challenge of heterogeneity in pathology image segmentation. By implementing a sophisticated three-tier alignment strategy—encompassing image-level collaborative style enhancement, feature-level adaptive feature alignment, and model-level stratified similarity aggregation—PathFL effectively mitigates representation bias and enhances cross-center generalizability. The framework systematically tackles complex heterogeneity across four heterogeneous datasets of sources, modalities, organs, and scanners, demonstrating superior segmentation accuracy and robustness. Our PathFL significantly outperforms the baseline, with average Dice improvements of 1.65\% for cross-source datasets, 6.13\% for cross-modality datasets, 7.94\% for cross-organ datasets, and 6.35\% for cross-scanner datasets. Experimental validation across four heterogeneous datasets substantiates PathFL's exceptional performance, highlighting its transformative potential for federated learning in medical image analysis. Future research directions include exploring PathFL's adaptability to additional medical imaging domains and practical medical implementation. Overall, this work lays a solid foundation for future research in federated learning and pathology image segmentation, paving the way for more effective and equitable applications in clinical diagnosis.

\section*{{Acknowledgments}}
{This work was supported by the National Natural Science Foundation of China (Grant No. 82441021), and the Big Data Computing Center of Southeast University for facility support. Additional support was provided by A*STAR Central Research Fund (“Robust and Trustworthy AI system for Multi-modality Healthcare”), and RIE2025 Industry Alignment Fund – Industry Collaboration Project (IAF-ICP) (Award No: I2301E0020).}

\printcredits

{\section*{Declaration of Competing Interest}
The authors declare that they have no known competing financial interests or personal relationships that could have appeared to influence the work reported in this paper.}

\bibliographystyle{cas-model2-names}

\bibliography{cas-refs}



\end{document}